%% file: main.tex
\documentclass[a4paper,12pt]{article}
\pdfoutput=1    

\usepackage[]{hyperref} 

\usepackage[utf8]{inputenc}
\usepackage{amsmath}
\usepackage{amsthm}
\usepackage{amssymb}
\usepackage{mathrsfs}
\usepackage{amsfonts}
\usepackage{bbold}

\usepackage{algorithm}
\usepackage{algorithmicx}
\usepackage{algpseudocode}
\usepackage{graphicx}
\usepackage{xifthen}
\usepackage{rotating}
\usepackage{xcolor}
\usepackage{subfig} 
\usepackage{psfrag}
\usepackage{amsopn}

\usepackage{authblk}

\usepackage{mathptmx}       
\usepackage{helvet}         
\usepackage{courier}        
\usepackage{type1cm}        
\usepackage{makeidx}         
\usepackage{graphicx}        
\usepackage{multicol}        
\usepackage[bottom]{footmisc}

\usepackage{ucs}              

\usepackage{latexsym}
\usepackage{amsmath,amssymb,amscd,amsbsy}
\usepackage{upgreek}

\usepackage{tensor}

\usepackage{makeidx}         
\usepackage{graphicx}        
\usepackage{multicol}        
\usepackage[bottom]{footmisc}

\usepackage{rcs} \RCS $Revision: 2.2 $ \RCS $Date: 2018/09/02 11:16:02 $ \RCS $Author: hgm $ \RCS $RCSfile: 18_RV-algebra-model.tex,v $ \RCS $Id: 18_RV-algebra-model.tex,v 2.2 2018/09/02 11:16:02 hgm Exp $

\usepackage{lmodern}   
\usepackage{anyfontsize}

\usepackage{mathrsfs}
\usepackage{bbm}
\input{components/settings-commands}
\input{components/iformula}

\usepackage[sfdefault=cmbr,OMLmathsans]{isomath}  
\usepackage{latexsym}
\usepackage{amsmath}
\usepackage{amsthm}
\usepackage{amssymb}
\usepackage{upgreek}  
\usepackage{tensor}  

\usepackage[toc,title,page]{appendix}

\usepackage{dcolumn}
\usepackage{tikz}
\usetikzlibrary{shapes,arrows}
\usetikzlibrary{intersections}

\usepackage{float}
\usepackage{xcolor}

\makeindex             

\begin{document}

\title{Convolutional generative adversarial imputation networks for spatio-temporal missing data in storm surge simulations}

\author{{\small Ehsan Adeli$^{\dag}$, Jize Zhang$^{\ddag}$, Alexandros A. Taflanidis$^{\dag}$}}
\affil{{\small $^{\dag}$ Department of Civil $\&$ Environmental Engineering $\&$ Earth Sciences\\ 
University of Notre Dame, Notre Dame, IN 46556, USA \\
$^{\ddag}$ Center for Applied Scientific Computing, Computing Division\\
Lawrence Livermore National Laboratories, Livermore, CA 94550, USA\\ 
eadeli@nd.edu}}
\date{July 2021}

\maketitle

\abstract{Imputation of missing data is a task that plays a vital role in a number of engineering and science applications. Often such missing data arise in experimental observations from limitations of sensors or post-processing transformation errors. Other times they arise from numerical and algorithmic constraints in computer simulations. One such instance and the application emphasis of this paper are numerical simulations of storm surge.   
The simulation data corresponds to time-series surge predictions over a number of save points within the geographic domain of interest, creating a spatio-temporal imputation problem where the surge points are heavily correlated spatially and temporally, and the missing values regions are structurally distributed at random.
Very recently, machine learning techniques such as neural network methods 
have been developed and employed for missing data imputation tasks.
Generative Adversarial Nets (GANs) and GAN-based techniques have particularly attracted attention as unsupervised machine learning methods. 
These models are trained to complete the missing data according to the underlying patterns. In this study, the Generative Adversarial Imputation Nets (GAIN) performance is improved by applying convolutional neural networks instead of fully connected layers to better capture the correlation of data and promote learning from the adjacent surge points. 
Another adjustment to the method needed specifically for the studied data is to consider the coordinates of the points as additional features to provide the model more information through the convolutional layers.
We name our proposed method as Convolutional Generative Adversarial Imputation Nets (Conv-GAIN).  
The proposed method's performance by considering the improvements and adaptations required for the storm surge data is assessed and compared to the original GAIN and a few other techniques. The results show that Conv-GAIN has better performance than the alternative methods on the studied data. 

\section{Introduction}
\label{sec:Introduction}
Challenges associated with missing data are frequent in a number of engineering and sciences applications \cite{Nakai, Kreindler, Ahmed, Rodrigues, Imani}.  Frequently such instances of missing data occur in experimental or survey observations for a variety of reasons, such as sensor failure, post-processing transformation errors, or financial constraints and experimental limitations that do not allow complete set of the desired features to be collected. When such incomplete datasets need to be exploited through techniques that assume availability of a complete data set, a necessary first step is the imputation of the data, to fill in the missing values. Once the complete data set is prepared, further observations, training, and learning based on the entire data can be established with no limitations. Beyond experimental observations, missing data may arise in computer simulations from numerical or physical constraints. One such example, and the application emphasis of this paper is storm surge simulations.\\

Numerical estimation of storm surge has been receiving increased attention the past decades \cite{Resio, Tanaka}, recognizing its potential contributions in improving resilience of coastal communities in a changing climate \cite{Anarde, Pant}. The predictions are useful for both real-time forecasting during landfilling events as well as for regional long-term risk assessment and planning support. To facilitate high-accuracy surge predictions, high-fidelity numerical models are typically used in coastal studies \cite{Kennedy2012, Nadal2015} to allow a detailed representation of the underlying storm hydrodynamic processes. Since such models have a substantial computational burden, surrogate modeling techniques are typically leveraged \cite{Irish, Gaofeng} to accommodate their further use in real-time surge forecasting and regional planning \cite{Kijewski, Nadal2020}. The development of surrogate models in this context requires an imputation of the numerical simulation data \cite{Jia, Shisler, Kyprioti}, the latter corresponding to peak-surge or surge time-series for a number of nodes of interest in the examined geographical domain (frequently referenced as save-points). The imputation pertains specifically to near-shore nodes that have remained dry in some of the storm simulations, and provides the so-called pseudosurge that completes the original database and allows further manipulation using standard machine learning and surrogate modeling techniques. The standard approach to perform the imputation \cite{Jia} is for each storm simulation, which leads to a spatial (for peak surge predictions) or spatio-temporal (for surge time-series predictions) imputation problem. So far, emphasis in the literature has been primarily given on imputation for peak-surge estimates \cite{Kijewski, Nadal2020}. \\

Examining, more broadly, data imputation techniques, initial attempts simply replaced missing data with global statistics \cite{Lin, Yaghoubi, Murray}, though recent efforts are exploring probabilistic and machine learning methods to learn from the existing observed patterns in the incomplete data. Examples include k-nearest neighbor (KNN) methods \cite{Jonsson, Batista}, Support Vector Machine applications (SVN) \cite{Pelckmansa}, Matrix completion and factorization \cite{Jiang, Wu, Chen} and MissForest \cite{Zhang, Stekhoven} approaches, Principal component analysis (PCA) \cite{Grung, Podani, Severson}, Kriging-based \cite{HYang, Bae, Song} or Gaussian Process (GP) \cite{Rodrigues, Imani} methods. The latter family (Kriging and GP) are particularly attractive for spatio-temporal problems, like the one considered here, though they might face few important challenges:  a) to efficiently handle large datasets (many nodes and many time instances) some covariance approximation/simplification will be needed \cite{Furrer, Genton, Cressie} that might reduce predictive accuracy; b) approach assumes correlation of surge between all nodes in close distance to one-another, which might not be the case for all near-shore coastal regions, since complex local geomorphologies (for example existence of barriers or riverine systems) might change the storm inundation characteristics even for nodes in geographic close proximity; c) missing data for storm surge imputation is not randomly distributed in space and time, rather it appears in structured format as will be shown later, with substantial part of nodes in the same geographical domain remaining dry for same time period, providing challenges in the calibration (proper selection of length and temporal correlation scales). Similar challenges exist for the other approaches referenced earlier that could be considered for data imputation in this setting.\\ 
 
An alternative formulation is examined here, based on Generative Adversarial Networks (GANs) \cite{Goodfellow, Yoon, Li, Kachuee, Kazemi, Cai, Shang, Luo}. Approach relies on learning the data spatially and temporally by means of a GAN-based model in order to observe the existing patterns in the corrupted dataset and therefore impute the missing components distributed structurally at random in the examined application as mentioned earlier. 
To overcome limitations for exploring the data correlations considering the studied data, we propose Conv-GAIN, a novel GAIN-based technique considering the spatial and temporal correlation of data, 
further taking advantage of node coordinates to enrich the provided information to the neural network model. To the best of our knowledge, our proposed Conv-GAIN is the first time that CNNs are coupled with the GAIN technique and applied for data imputation problems. Our proposed Conv-GAIN can capture both spatio-temporal features simultaneously.
Though advances are examined within the specific storm surge imputation task that motivates this research, they have broader applicability for other problems that share the same underlying features, with correlation characteristics that can be learned by the neural network model to assist the process of imputation for different types of missing data distribution.\\
  
The remainder of the paper is organized as follows. Section \ref{section:Storm Surge Imputation Problem Characteristics} examines the problem set up for the specific application of interest, while Section \ref{section:Neural Network Imputation Methods} provides the methodological developments for the proposed neural network imputation. Section \ref{section:Illustrative Application} presents the data used for validation of the method and 
provides results as well as comparisons to alternative formulations (mainly PCA) that could be applied to the given dataset and Section \ref{section:Conclusion} summarizes the observations and conclusions of this study. \\

\section{Storm Surge Imputation Problem Characteristics}
\label{section:Storm Surge Imputation Problem Characteristics}
Storm surge numerical simulations provide predictions for the storm surge $\upsilon$ as a function of time $t \in \mathbb{R}$ and space $s \in \mathbb{R}^2$ defined through latitude and longitude coordinates. Imputation is considered here separately for each storm, though evidently this needs to be repeated for each storm in the available dataset. For time-series predictions, data typically refers to specific save-points (SP) within the geographic domain of interest, corresponding to a subset of the computational grid utilized in the high-fidelity model \cite{Jia}. This leads to data for each storm that is discretized in time and space $\mat{\Upsilon}_{ij}, i = 1, .., n_t, j = 1, .., n_s$, with $n_t$ denoting the total number of time-instances and $n_s$ the total number of nodes. The corresponding matrix of discretized values will be denoted herein as $\mat{\Upsilon}$ with rows corresponding to the different SP (nodes) and columns to the different times. Discretization in time typically corresponds to equally spaced intervals. On the other hand, the SPs correspond to an irregular grid with varying density within the geographic domain of interest. Figure \ref{fig:stormgrid} shows  an example for 
SP distribution, using the database that will be later used in the illustrative case study. 

\begin{figure}[H]
    \begin{center}{}
      \includegraphics[width=3.5in]{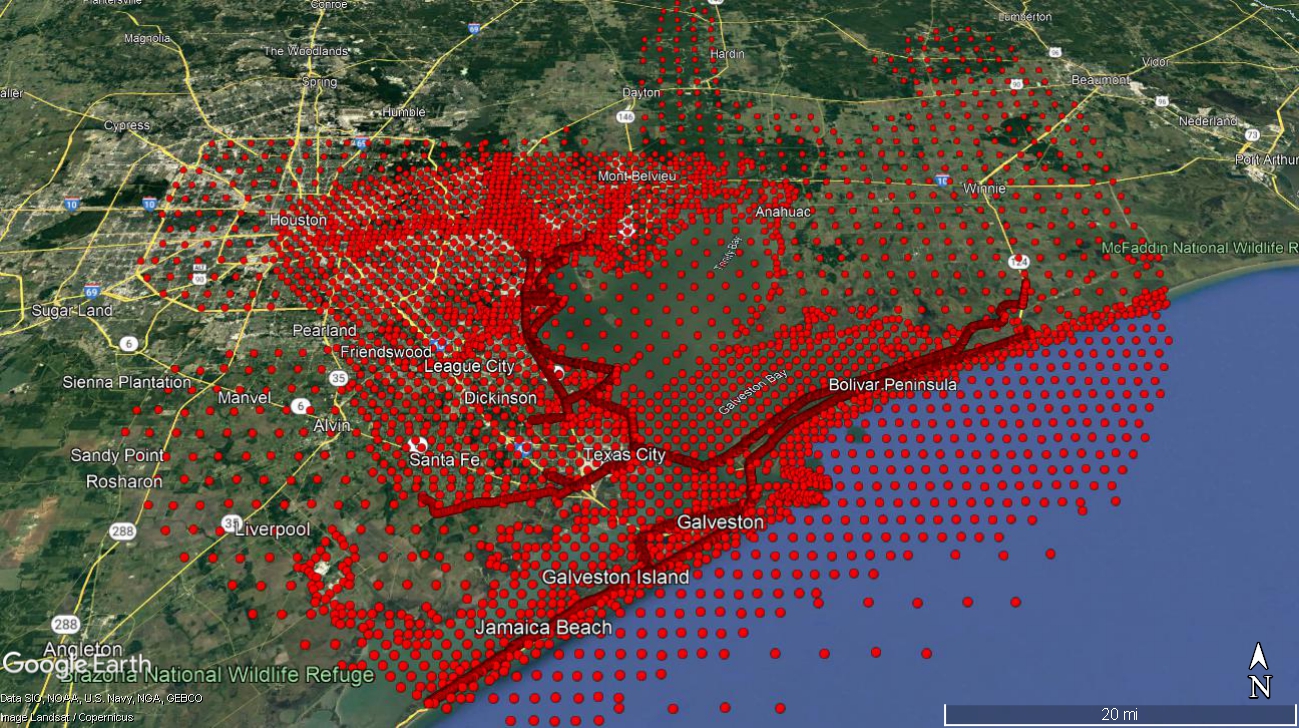}
      \caption{\label{fig:stormgrid} Example of save point (SP) distribution from a typical coastal flood study. Red dots indicate save point (SP) locations.}
    \end{center}\hspace{2pc}%
\end{figure} 

This irregularity stems from the fact that the original numerical grid is itself irregular and the fact that higher density of SP is needed close to shore, where surge is expected to have greater variability (nonlinear phenomena) \cite{Resio}. Note that for describing the temporal variation of the storm surge, distance of eye of the storm from landfall can be used instead of time to landfall, but to maintain consistency in terminology with the broader literature on spatio-temporal imputation, variation will be described as referring to time $(t)$ herein. \\

Missing data in this case corresponds to time instances for which SP are dry; these might pertain to onshore nodes that are originally dry and get inundated as the storm surge evolves, or to nodes that have been originally wet but become dry during some time period due to so-called negative surge effect when water mass drains from parts of the geographic domain and moves to others as the storm approaches landfall. 
Imputation is required to complete the predictions and estimate the pseudo-surge for such instances. The corresponding imputation problem has the following attributes. Values for $n_t$ and especially $n_s$ can be very large creating a potential computational complexity. Spatial ordering of the SPs (rows of matrix $\mat{\Upsilon}$) is impractical since data corresponds to an irregular two-dimensional grid. Note that the same does not apply to temporal data, which are ordered and regularly spaced. Missing data are not structurally distributed at random within matrix $\mat{\Upsilon}$ 
as many existing imputation methodologies assume \cite{HYang, Bae, Imani}, rather they follow certain patterns, with SP remaining consecutively dry for multiple time instances 
and nodes in close-proximity frequently becoming dry simultaneously. Finally, due to complex local geomorphologies, 
such as existence of disjointed near-shore 
riverine systems or protective barriers, variation of surge can appear to be quite discontinuous in certain parts of the geographic domain (though in general a smooth variation is expected). Latter characteristic imposes problems for imputation techniques that rely exclusively on correlation of nodes based on distance. Though such correlation information can be useful, imputation using this information exclusively might provide erroneous information for any domains that fall in the aforementioned categories with complex local geomorphologies. To address these challenges a neural network approach is examined in the next section for spatio-temporal imputation.

\section{Neural Network Imputation Methods}
\label{section:Neural Network Imputation Methods}

\subsection{Generative Adversarial Networks}
\label{subsection:Generative Adversarial Networks}

Generative Adversarial Networks (GANs) \cite{Goodfellow} are a group of unsupervised machine learning techniques that leverages a game-theoretic framework to extract the implicit distribution of the input data. This is performed by adversarial networks where a discriminator network estimates the probability of a data instance being real or fake. The data instances are mapped from a latent variable z by a generative model called generator which is trained to deceive the discriminator. The discriminator is therefore trained in a supervised way, where real and fake data are labeled as one and zero, respectively \cite{Goodfellow, Vanilla, Luc, Mescheder, Ho}.\\

GANs consist of two sub-models. The first model is the generative model that tries to learn from the dataset by capturing the data distribution to generate new plausible examples from the problem domain while training. The second model is the discriminator model that estimates the probability that a sample comes from the training data rather than the generator \cite{Goodfellow, Mescheder}. In fact, it tries to categorize the examples into two groups; the real examples, which are from the domain, and the fake examples, which are in reality generated by the generator model. These two models are trained simultaneously until the generator model generates plausible examples, and it happens when the discriminator model is fooled properly. In other words, the generator outputs a batch of samples and provides them to the discriminator along with real examples from the domain. The discriminator then tries to distinguish the real and fake examples, so it gets better at discriminating in the next round \cite{Goodfellow, Vanilla}. \\  

The generator's performance depends on how well the fake samples fooled the discriminator, while the generator learns how to create samples drawn from the same distribution as the training data. Strictly speaking, the training procedure for generators is to maximize the probability of the discriminator making a mistake \cite{Mescheder, Vanilla}. The zero-sum is here defined where the generator is penalized with significant updates to model parameters. At the same time, no change is needed for the discriminator when it successfully determines the fake and real samples \cite{Goodfellow, Karpathy}. GANs can be used for a wide range of applications, most notably in image translations tasks \cite{Caesar}. Recently there has been significant attention toward the application of GANs because of their success in different areas. In the next section, we discuss several techniques based on modifications of GANs applied for the imputation of missing data.\\ 

\subsection{GAN-based Imputation Techniques}
\label{subsection:GANs-based Imputation Techniques}

Generalizing approach, let us assume $\mat{X} \in \mathbb{R}^{n_t \times n_s}$ is a complete data matrix and $\mat{M}$ is a binary mask where $\mat{M} \in \{0, 1\}^{n_t} \times \{0, 1\}^{n_s}$ determines which entries in $\mat{X}$ to reveal $\mat{X} \sim \mathbb{P}_{\vek{\theta}}(\mat{X})$ and $\mat{M} \sim \mathbb{P}_{\vek{\phi}}(\mat{M} | \mat{X})$. Here $\vek{\theta}$ is the unknown parameters of the data distribution and $\vek{\phi}$ represents the unknown parameters of the mask distribution \cite{Rubin, Little}. If we denote the observed and missing elements of the vector $\mat{X}$ with $\mat{X}_o$ and $\mat{X}_m$ respectively, the missing data mechanism $\mathbb{P}_{\vek{\phi}} (\mat{M} | \mat{X}_{o}, \mat{X}_{m})$ can be categorized in terms of independence relations between the complete data $\mat{X}$ and the masks $\mat{M}$ in three different classes \cite{Rubin, Little} explained below.\\

The first class is when the data are missing completely at random (MCAR). In this case $\mathbb{P}_{\vek{\phi}}(\mat{M} | \mat{X} ) = \mathbb{P}_{\vek{\phi}} (\mat{M})$ as the missingness occurs entirely at random. There is therefore no dependency of $\mathbb{P}_{\vek{\phi}}(\mat{M} | \mat{X} )$ on any of the variables. In the second class, the data is missing at random (MAR) and the missingness depends only on the observed variables. In this case, $\mathbb{P}_{\vek{\phi}}(\mat{M} | \mat{X} ) = \mathbb{P}_{\vek{\phi}} (\mat{M} | \mat{X}_o)$. In the third class, the data is missing not at random (NMAR) and the missingness is neither MCAR nor MAR. In this case, $\mat{M}$ depends on $\mat{X}_m$ and possibly also on $\mat{X}_m$ \cite{Rubin}. It should be pointed out that in the first and second categories the $\mathbb{P}(\mat{X}_o, \mat{M})$ can be factorized into $\mathbb{P}_{\vek{\theta}} (\mat{X}_o) \mathbb{P}_{\vek{\phi}}(\mat{M} | \mat{X}_o)$ \cite{Little}. Accordingly, most work on the imputation of missing data for the sake of simplicity assumes that the corrupted data are either MAR or MCAR. \\

There are a limited number of research works where the missing data are imputed by applying GAN techniques. A generative adversarial network method to impute the missing data, which are entirely at random, is introduced by Yoon et al. \cite{Yoon}. They proposed a GAN-based method and investigated the efficiency of this method by comparing its performance with some other probabilistic and machine learning techniques for different sets of corrupted clinical data. Li et al. \cite{Li} has proposed using several discriminator and generator models in order to learn the patterns in the missing data and data distribution to complete the corrupted data. 
The performance of this work is later investigated and compared to a generative imputation technique in a research paper by Kachuee et al. \cite{Kachuee} for picture datasets. They also applied and tested their proposed approach for different types of datasets. Kazemi et al. \cite{Kazemi} has recently introduced an iterative generative adversarial network and applied it to experimental traffic data. Cai et al. \cite{Cai}, Shang et al. \cite{Shang}, and Luo et al. \cite{Luo} have also introduced some GAN-based methods to impute the missing data for clinical applications and other applications and compared the performance of their approaches with some stochastic methods as well as the research studies and methods mentioned above.\\

In most of the discussed approaches above, the missing subregions in the corrupted data are distributed either randomly or entirely randomly. 
As discussed in Section 2, in the application examined here, the experimental corrupted data contain missing regions which are distributed not randomly but structurally at random. In other words, the missing areas are the mixtures of huge blocks 
distributed with no apparent patterns. To consider the information from the spatial and temporal correlation of data, and also to accommodate for the structurally distribution of missing data 
the technique introduced by Yoon et al. \cite{Yoon} is adopted and 
further advanced based on the concept of convolutional neural networks (CNNs) \cite{LeCun}.

\subsection{Convolutional Generative Adversarial Imputation Nets (Conv-GAINs)}
\label{subsection:Convolutional Generative Adversarial Imputation Nets (Conv-GAINs)}

Following \ref{subsection:GANs-based Imputation Techniques}, 
$\mat{X}_m$ denotes the matrix representing missing (unobserved) components. 
In the context of the surge imputation problem presented in Section \ref{section:Storm Surge Imputation Problem Characteristics}, $\mat{X}_m$ corresponds to the missing values of $\mat{\Upsilon}$, with  $\mat{X}_o$ corresponding to surge values for all time-instances each node has been inundated.
The objective is to impute $\mat{X}_m$ 
by generating samples from the conditional distribution of $\mat{X}$ given $\mat{X}_m$, $\mathbb{P}(\mat{X} | \mat{X}_m)$, to complete the missing data components in our dataset. \\

The introduced Conv-GAIN is inspired by GAIN \cite{Yoon}, so let us first revisit the fundamental of GAIN. The modifications and improvements for the proposed spatio-temporal Conv-GAIN are also noted in this section. In the original GAIN, the generator takes $\mat{X}_m$, $\mat{M}$, and $\mat{Z}$ where the latter is noise variables, independent of all other variables, and outputs 
$\mat{V}$ representing the imputation values. A generator imputer $G(\mat{X}, \mat{M}, \mat{Z}) = (\mat{U}, \mat{V})$ is accordingly calculated as follows.

\begin{subequations}
\label{generator}
\begin{align}
& \mat{U} = \mat{X} \circ \mat{M} + \mat{Z} \circ (\mat{1}-\mat{M}) 
\\ 
& \mat{V} = \mat{X} \circ \mat{M} + \ops{g}(\mat{U}) \circ (\mat{1}-\mat{M}),  
\end{align}
\end{subequations}

\noindent where Hadamard production \cite{Neudecker} is denoted by $\circ$, function $\ops{g}(.)$ is to be learned, $\mat{U}$ is the intermediate imputed values for all observed and unobserved components, and $\mat{V}$ corresponds to the final imputed components. Note that the observed components are in the final step replaced in $\mat{V}$, i.e. $\mat{V} = \mat{X}_o$ for the similar components indexes. 
It should be pointed out that the generator structure in this neural network model is similar to the generator structure in any standard GANs \cite{Yoon}. \\

The discriminator in the model determines which components in the imputed values provided by the generator are fake, which means they are imputed by the generator, or real, which means they are observed values \cite{Yoon}. In contrast to GAN, that the entire data values are identified as either fake or real, here this determination is for each single component \cite{Yoon, Goodfellow}. Therefore, the discriminator maximizes the probability of predicting mask $\mat{M}$ based on the imputed data by the generator $\mat{V}$. The hint mechanism in GAIN is also introduced to control the amount of information carries by the hint matrix $\mat{H}$ about the $\mat{M}$. Hint matrix $\mat{H}$ is obtained through the equation $\mat{H} = \mat{B} \circ \mat{M} + 0.5(1- \mat{B}) $, where $\mat{B}$ is a random variable matrix with similar shape with $\mat{M}$, which is defined by sampling $k$ uniformly at random, and then setting the components of $\mat{B}$ equal to zero for the similar indices of samples and $\mat{B}$ equal to one for the rest of the components. In case we do not provide enough information about $\mat{M}$ through $\mat{H}$, there could be several distributions that the generator could provide, and all of them would be optimal with respect to the discriminator \cite{Yoon}.
A schematic representation of GAIN is shown in Figure \ref{fig:GAINarchitecture} where the generator and discriminator are indicated by G and D, respectively.

\begin{figure}[H]
    \begin{center}{}
      \includegraphics[width=3.5in]{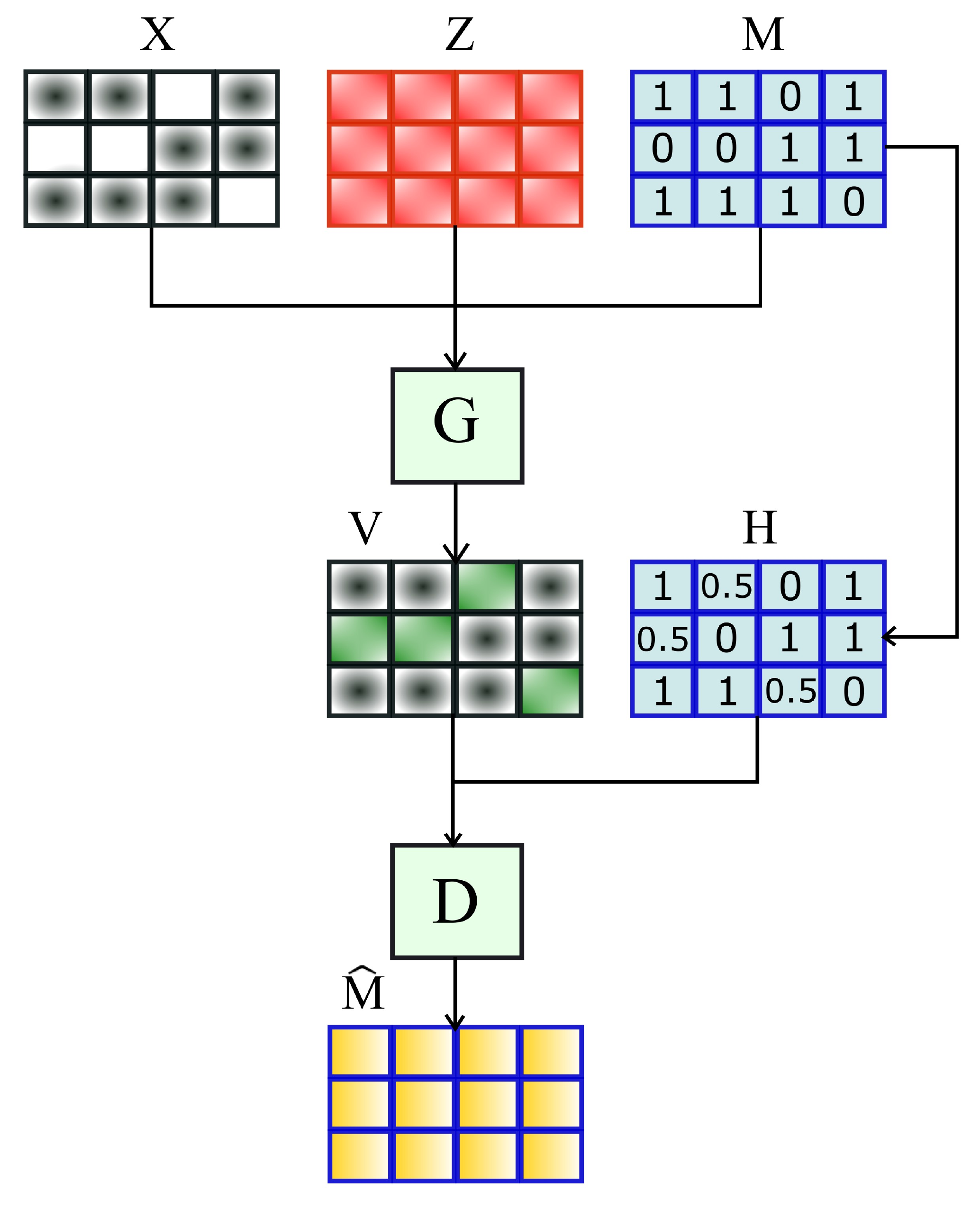}
      \caption{\label{fig:GAINarchitecture} Conv-GAIN architecture}
    \end{center}\hspace{2pc}%
\end{figure} 

Introducing the output of the discriminator as $\hat{\mat{M}}$ whose distance from $\mat{M}$ is minimized, the cost function of the generator and discriminator are respectively defined as follows.

\begin{equation}
\mathbb{J}^{(G)}( \vek{\theta}^{(G)}, \vek{\theta}^{(D)}) = - 
\mathbb{E}_{\mat{M}, \hat{\mat{M}}} [ \Sigma_{i  =1}^{n_t} \Sigma_{j  =1}^{n_s} [(\mat{1}-\mat{M}_{ij}) \log (\hat{\mat{M}}_{ij}) - \alpha 
\ops{L_2} (\mat{V}_{ij} - \mat{X}_{ij})] 
\label{generatorlossfunction}
\end{equation} 

\begin{equation}
\mathbb{J}^{(D)}( \vek{\theta}^{(G)}, \vek{\theta}^{(D)}) = - \mathbb{E}_{\mat{M}, \hat{\mat{M}}} [ \Sigma_{i  =1}^{n_t} \Sigma_{j  =1}^{n_s} [ \mat{M}_{ij} \log(\hat{\mat{M}}_{ij})  +  (\mat{1}-\mat{M}_{ij}) \log (\mat{1} - \hat{\mat{M}}_{ij})  ]],
 \label{discriminatorlossfunction}
\end{equation}

\noindent where $\mathbb{J}$ and $\mathbb{E}$ stand for the cost function and expectation respectively,
$\log$ is the element-wise logarithm, $\ops{L_2}$-norm is the distance of the imputed values of the observed components from the actual observed components, and $\alpha$ is a hyper-parameter \cite{Yoon}.\\

$\log$ is the element-wise logarithm, $\ops{L_2}$-norm is the distance of the imputed values of the observed components from the actual observed components, and $\alpha$ is a hyper-parameter \cite{Yoon}.\\

The goal is thus to maximize the probability of correctly predicting $\mat{M}$ by training the discriminator. At the same time, while training the generator, we aim to minimize the probability of the discriminator predicting the $\mat{M}$. 
The objective of GAIN is defined to be the min-max problem given by

\begin{equation}
\underset{G}{\ops{min}} \,  \underset{D}{\ops{max}} \, \mathbb{J}^{(D)}(D, G),
 \label{gainglossfunction}
\end{equation} 

\noindent which should be first optimized while the generator is fixed. Once the discriminator is optimized, the generator should be optimized. The generator imputes the entire data. 
We, therefore, should not only ensure that the imputed values for the missing components fool the discriminator through the min-max game but also the values outputted by the generator for the observed components are meaningfully close to the actual observed components \cite{Yoon}. These two are fulfilled through the first and second terms of the formulation in Eq. \ref{generatorlossfunction}. The second term is, in fact, one more loss function which is therefore defined to minimize the distance of the components that are truly observed. Finally, the generator is optimized by minimizing the weighted sum of the two generator losses, Eq. \ref{generatorlossfunction}, through the min-max problem below.\\

\begin{equation}
\underset{G}{\ops{min}} \,  \underset{D}{\ops{max}} \, \mathbb{J}^{(G)}(D, G)
 \label{gaindlossfunction}
\end{equation}

However, a 
drawback of the original GAIN approach \cite{Yoon} is its use of fully connected neural networks in both discriminators and generators, leaving the spatial and temporal correlation of data unexplored. Therefore, some discriminative information cannot be extracted to improve the imputation performance. 
As described earlier, this motivates the proposed improvements, presented in detail next. \\

To overcome the limitations, we propose and evaluate a set of convolutional neural network layers
. The introduction of convolutional layers is aimed to capture the correlation of data temporally and spatially, so that the imputation of missing data can be conducted in a more informative manner. In addition, the convolutional layers enable the data imputation approach to learn from the adjacent nodes which is very appropriate for the
studied datasets with structurally distributed missing data. 
Through this approach, as the series of data in the temporal dimension and the node order dimension are observed by filters in convolutional layers, the corresponding temporal/intra-node correlations with their adjacent time steps and nodes would be taken
into account. We name this class of adversarial networks as Convolutional Generative
Adversarial Imputation Nets (Conv-GAINs).\\

In addition, to increase the amount of information we can provide to the convolution layers, we incorporate the spatial coordinates of the nodes as additional features along with their surge values as the input to the convolutional neural networks. With such treatment, the convolution operations will be conducted both spatially and temporally, and enables the neural networks to learn the spatio-temporal correlation simultaneously. The benefit of this step will be further discussed in Section \ref{subsection:Effect of Structural Distribution of Missing Data and Spatial Features}, where we will compare its imputation capability with another Conv-GAIN approach that does not use the spatial coordinates as additional features. \\

In the convolutional and dense layers of Conv-GAIN, ReLU \cite{Han} is used for the hidden layers except for the last layers as its performance is generally better than most of the other activation functions. For the last layers, sigmoid activation function is used that outputs between zero and one \cite{Han}.
Furthermore, max-pooling layers \cite{Dan} are applied to improve the performance of convolutional neural networks. Pooling layers by reducing the small translation of the features would decrease the total amount of parameters in convolution layer as it downsamples the volume of convolution neural network. Accordingly the computation cost of the network decreases. 
Adam optimizer is applied to optimize the minimization problems. Besides batch size, the other hyper-parameters like number of layers and hint rate are tuned through random search approach \cite{Bergstra}.\\

The order and type of the layers of GAIN and Conv-GAIN models designed and applied in this work are described in Table \ref{tab:layers}. The filter size for convolution and pooling layers are also denoted in this table 
where G and D stand for the generator and discriminator models. Dense layers are the fully connected layers which are used in GAIN with the output size mentioned for each dense layer. In Conv-GAIN model, right after each convolution layer a pooling layer is used to reduce the dimension. The filter size, $3\times3$ and number of channel for each convolution layer is denoted in Table \ref{tab:layers}. The filter size $2\times2$ for the pooling layer is also described. After convolutional and pooling layers, number of outputs are decreased gradually through two dense layers in the end of the neural network models so the models do not experience a dramatic change of parameter size.

\begin{table}[H]
\caption{Neural network model architectures, with filter size and number of channels for each convolutional and pooling layer reported}
\vspace{-0.5cm}
\begin{center}
\scalebox{1.0}{
\begin{tabular}{l*{3}{c}r }
\hline
Model      & Layer & Filter & Output &\\
\hline
  $\text{GAIN} $ \hspace{1cm} G: & $\text{Dense}$& $ $ & $\text{[125]}$ &\\
  $ $                       & $\text{Dense}$& $ $ & $\text{[125]}$ &\\
  $ $                       & $\text{Dense}$& $ $ & $\text{[125]}$ &\\
  $ $                       & $\text{Dense}$& $ $ & $\text{[125]}$ &\\
  $ $ \hspace{2.1cm} D: & $\text{Dense}$& $ $ & $\text{[125]}$ &\\
    $ $                       & $\text{Dense}$& $ $ & $\text{[125]}$ &\\
    $ $                       & $\text{Dense}$& $ $ & $\text{[125]}$ &\\
    $ $                       & $\text{Dense}$& $ $ & $\text{[125]}$ &\\
  $\text{Conv-GAIN}$ \hspace{0.1cm} G: & $\text{Convolution}$& $[3,3,32]$ & $\text{[6, 125, 32]}$ &\\
  $ $                       & $\text{Pooling}$& $[2, 2]$ & $\text{[3, 63, 32]}$ &\\
  $ $             &  $\text{Convolution}$& $[3,3,64]$ & $\text{[3, 63, 64]}$ & \\
  $ $                       & $\text{Pooling}$& $[2, 2]$ & $\text{[2, 32, 64]}$ &\\
  $ $ & $\text{Data flatten} $ & $ $ & $\text{[4096]}$ &\\
  $ $                       &  $\text{Dense}$& $ $ & $\text{[1024]}$ &\\  
  $ $                       & $\text{Dense}$& $ $ & $\text{[375]}$ &\\
  $ $ \hspace{2.12cm} D: & $\text{Convolution}$& $[3,3,32]$ & $\text{[6, 125, 32]}$ &\\
    $ $                       & $\text{Pooling}$& $[2, 2]$ & $\text{[3, 63, 32]}$ &\\
    $ $             &  $\text{Convolution}$& $[3,3,64]$ & $\text{[3, 63, 64]}$ & \\
    $ $                       & $\text{Pooling}$& $[2, 2]$ & $\text{[2, 32, 64]}$ &\\
    $ $ & $\text{Data flatten} $ & $ $ & $\text{[4096]}$ &\\
    $ $                       &  $\text{Dense}$& $ $ & $\text{[1024]}$ &\\  
    $ $                       & $\text{Dense}$& $ $ & $\text{[375]}$ &\\
\hline
\end{tabular}}\label{tab:layers}
\end{center}
\end{table}

The layers of the applied Conv-GAIN are schematically indicated in Figure \ref{fig:schematiclayer}. In this figure, the output data and filter sizes are plotted by scaling their real size data dimensions.  \\

\begin{figure}[H]
    \begin{center}{}
      \includegraphics[width=4.0in]{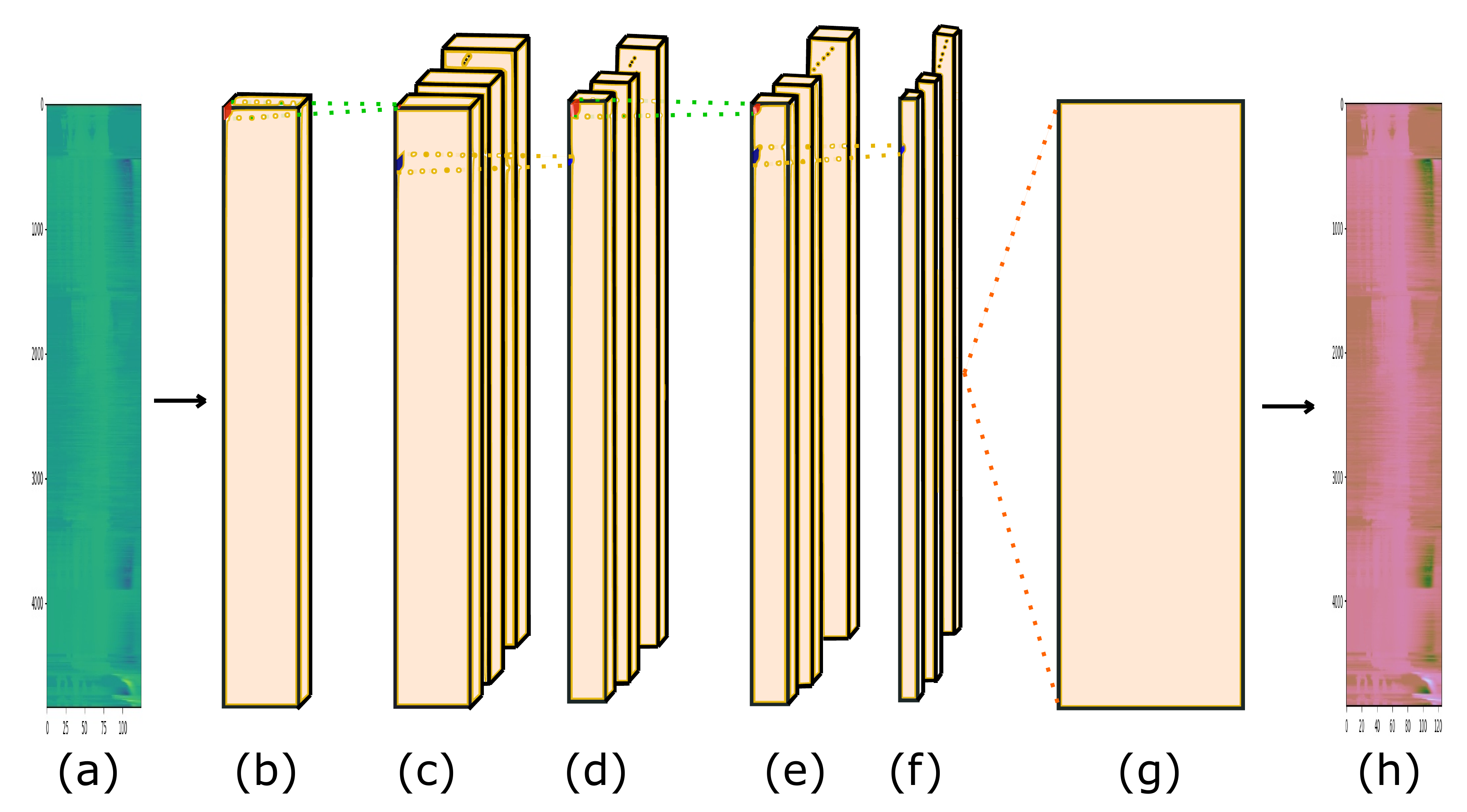}
      \caption{\label{fig:schematiclayer} Schematic structure of layers}
    \end{center}\hspace{2pc}%
\end{figure} 

In Figure \ref{fig:schematiclayer}, given the input data in part $\text{(a)}$, part $\text{(b)}$ shows the input data for the first convolutional layer where the coordinates of the nodes are also added as input features. Part $\text{(c)}$ shows then the output data of the first convolution layer with a filter size $3 \times 3$ with 32 channels. A pooling layer follows this convolutional layer shown in part $\text{(d)}$ with the filter size $2 \times 2$, which divides the data size by two in each dimension. Part $\text{(e)}$ indicates the output data of the second convolutional layer with a similar filter size with the first convolutional layer and 64 channels. After the second convolutional layer, the second pooling layer with an identical filter size to the first pooling layer is used. The output data of the second pooling layer is indicated in part $\text{(f)}$, where the size of data is again reduced. The data are then passed through two connected layers after they are flattened, and output data are shown in part $\text{(g)}$. Finally, the imputed data, part $\text{(h)}$, is derived from the last layer of the neural network model. Note that a similar set of layers is used for both the generative and discriminator models of the Conv-GAIN model.\\

The pseudo-code of the algorithm is also presented in Algorithm \ref{alg:Conv-GAIN}.

\begin{algorithm}[H]
\caption{Pseudo-code of Conv-GAIN}\label{alg:Conv-GAIN}
\begin{algorithmic}
\While{training loss has not converged}

\textbf{Discriminator optimization}
\State Draw $\ops{k_D}$ samples from datasets $\mat{X}$ and $\mat{M}$
\State Draw $\ops{k_D}$ i.i.d samples of $\mat{Z}$ and $\mat{B}$
\For{$\ops{k_D}$ samples}
\State  $\mat{U} \gets \mat{X} \circ \mat{M} + \mat{Z} \circ (\mat{1}-\mat{M})$ 
\State  $\mat{V} \gets \mat{X} \circ \mat{M} + \ops{g}(\mat{U}) \circ (\mat{1}-\mat{M})$
\State  $\mat{H} \gets \mat{B} \circ \mat{M} + 0.5(1- \mat{B})$
\EndFor
\State \text{Update D using stochastic gradient descent}

\textbf{Generator optimization}
\State Draw $\ops{k_G}$ samples from datasets $\mat{X}$ and $\mat{M}$
\State Draw $\ops{k_G}$ i.i.d samples of $\mat{Z}$ and $\mat{B}$
\For{$\ops{k_G}$ samples}
\State  $\mat{H} \gets \mat{B} \circ \mat{M} + 0.5(1- \mat{B})$
\EndFor
\State \text{Update G using stochastic gradient descent for fixed D}
\EndWhile
\end{algorithmic}
\end{algorithm}

\section{Illustrative Application}
\label{section:Illustrative Application}
This section looks at the results obtained by Conv-GAIN applied to a series of storm surge data. 
In the following sections, we first describe the desired data and discuss the results obtained by the mentioned methods.

\subsection{Dataset Description}
\label{subsection:Dataset Description}

The dataset utilized for the illustrative application corresponds to time-series surge predictions for the coastal Texas geographic region. Data is provided through the Coastal Hazards System (CHS) of the Army Corps of Engineers \cite{Nadal20201}. A total of 30 different storms are examined here. To accommodate a test-sample validation, only save points with originally complete information are considered for each storm, and structured missing values are created by adjusting the node elevation of the save points through the following process. 
The node elevation is gradually increased across the geographic domain, and save points for which their adjusted node elevation is smaller than their recorded surge are classified to correspond to missing values. Different elevation adjustments are chosen so that storms with different rates of missing data are obtained, corresponding to rates of $5\%$, $10\%$, $15\%$, $20\%$, $25\%$ and $30\%$. These correspond to the rates of missing data for the actual, original database, 
and allow us to explore the imputation performance across storms with different missing data characteristics, and different associated challenges. Five storms are chosen in each of these six ranges, yielding the total of 30 storms considered. 
It is important to stress that the specific development of the test-sample 
facilitates the same structure for the missing data as in the original database. This is verified in Appendix \ref{Dataset Characteristics}, showing surge heat-maps for the 30 storms. The structured pattern of the missing data is evident 
in the plots provided there.

\subsection{Imputation Performance Evaluation}
\label{subsection:Imputation Performance Evaluation}
The performance of Conv-GAIN is assessed across the storms with different missing rates in comparison to another GAN-based approach, GAIN, and two popular imputation baseline methods, principal component analysis (PCA), and 
Mean Imputation (MI). The experiment is conducted five times for each dataset, and the average error is calculated. The root square mean error (RMSE) is utilized as validation metric, calculated by comparing imputed surge values to the original surge values (ground truth). The RMSE is reported in Table \ref{tab:RMSE} and plotted in Figure \ref{fig:RMSE} for the group of storms with the different percentages of missing data.

\begin{table}[H]
\caption{RMSE of missing data}
\vspace{-0.5cm}
\begin{center}
\scalebox{0.8}{
\begin{tabular}{l*{7}{c}r }
\hline
Percentage      & $5\%$  &  $10\%$ & $15\%$ & $20\%$ &$25\%$ & $30\%$ &\\
\hline
  $\text{MI}$ & $2.230\mathrm{e}-1$ & $2.089\mathrm{e}-1$ & $3.213\mathrm{e}-1$ & $3.734\mathrm{e}-1$ & $3.500\mathrm{e}-1$ & $6.191\mathrm{e}-1$ &  \\
  $\text{PCA}$ & $9.178\mathrm{e}-2$ & $1.302\mathrm{e}-1$ & $1.562\mathrm{e}-1$ & $1.949\mathrm{e}-1$ & $2.292\mathrm{e}-1$ & $3.900\mathrm{e}-1$ &  \\
  $\text{GAIN}$             & $6.888\mathrm{e}-2$   & $7.118\mathrm{e}-2$ & $8.814\mathrm{e}-2$ &$1.381\mathrm{e}-1$ & $1.702\mathrm{e}-1$ & $2.776\mathrm{e}-1$ &\\
  $\text{\bf{Conv-GAIN}}$                  & $\bf{5.494\mathrm{e}-2}$ & $\bf{5.791\mathrm{e}-2}$ & $\bf{7.026\mathrm{e}-2}$ &  $ \bf{9.928\mathrm{e}-2}$ & $\bf{1.214\mathrm{e}-1}$  & $\bf{2.107\mathrm{e}-1}$  & \\
\hline
\end{tabular}}\label{tab:RMSE}
\end{center}
\end{table}

\begin{figure}[H]
    \begin{center}{} 
      \includegraphics[width=2.75in]{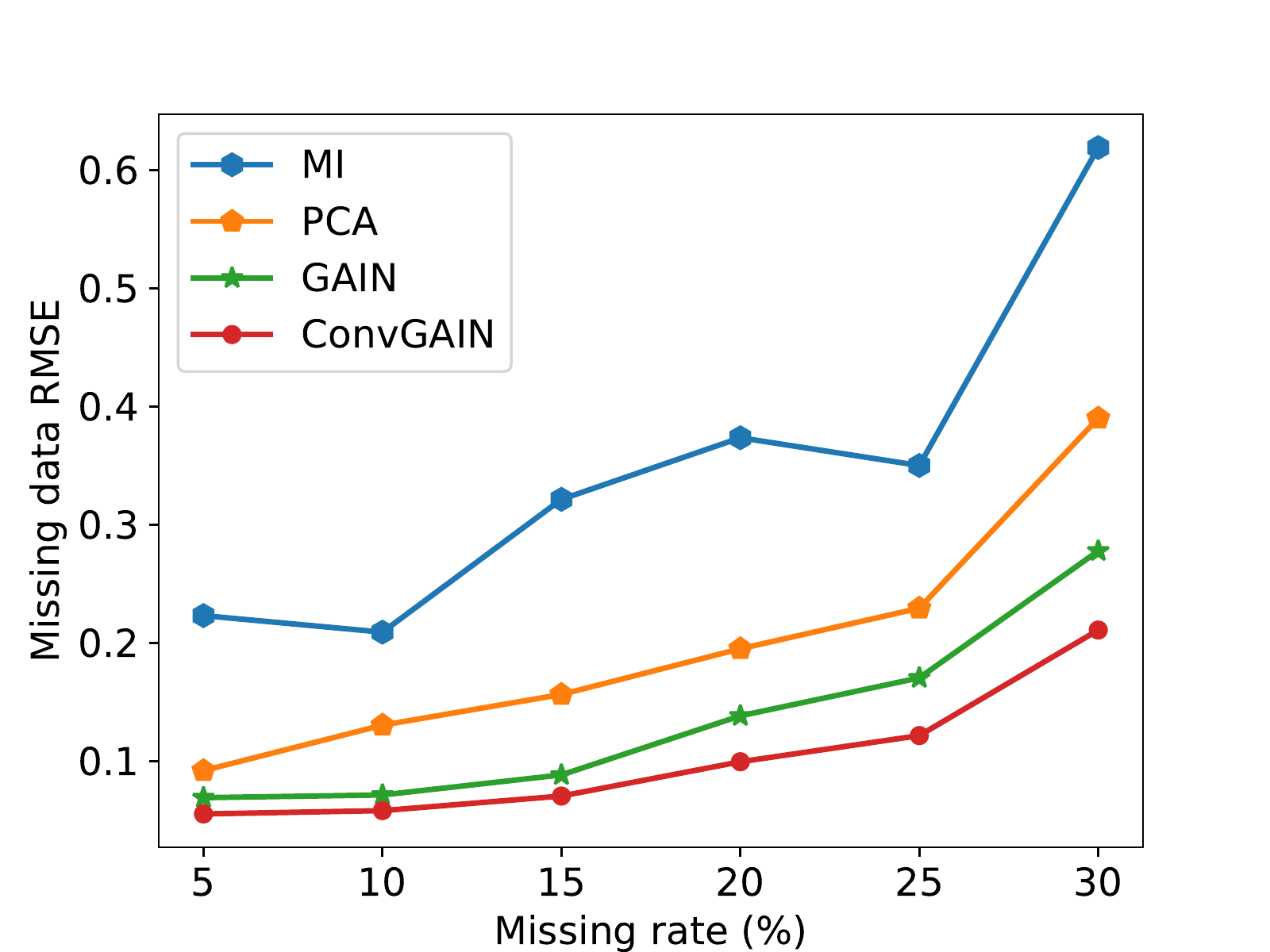}
      \caption{\label{fig:RMSE} RMSE performance of different methods}
    \end{center}\hspace{2pc}%
\end{figure} 

As it is seen in Figure \ref{fig:RMSE}, for the very low missing rates, the performance of GAIN and Conv-GAIN are very similar. For such datasets we generally do not observe dense corrupted missing data subregions as the missing data (see Appendix \ref{Dataset Characteristics})  is distributed more randomly and 
representing a less challenging task to conduct.
PCA performance also performs relatively well for very low missing rate (e.g., $5 \%$), but poorly for other datasets with higher missing rates. MI performance has also a meaningful gap with the other deployed approaches, something expected as it establishes a very simplistic imputation process.\\ 

As the rate of missing data in the datasets increases, the error of all approaches increases. 
The error of Conv-GAIN only gradually increases from $5.494\mathrm{e}-2$ to $2.107\mathrm{e}-1$ as the missing rate increases from $5 \%$ to $30 \%$, making Conv-GAIN the best performer across all the missing rates.
One possible explanation for the better  performance of Conv-GAIN is that it can learn from the neighboring' components by capturing the spatial and temporal correlation of data. Accordingly, for the datasets with multiple dense blocks of missing data and therefore higher missing rate, the imputation job can be carried better by Conv-GAIN than GAIN, 
since the latter cannot learn as efficiently from the adjacent regions to extract information. 
Both PCA and MI results also show that their performance are way worse 
on storm surge data with high missing rate than the two applied GAN-based methods. \\ 

To further examine the temporal consistency of the imputed data, we select four nodes 
from one examined storm dataset with a total missing rate of 
$30 \%$, and present how the imputation evolves over time. 
The missing rates for each of these four nodes is $36 \%$, $48 \%$, $60 \%$, and $72 \%$, respectively. The performance of the applied methods is shown in Figure \ref{fig:onenodeimputation}. Note that in this figure, the observed data represents the assumed available data for the imputation (i.e., elements that are not missing) 
while the reference values stand for the ground-truth values of the missing surge elements from the provided database.

\begin{figure}[H]
    \begin{center}{}
      \includegraphics[width=2.5in]{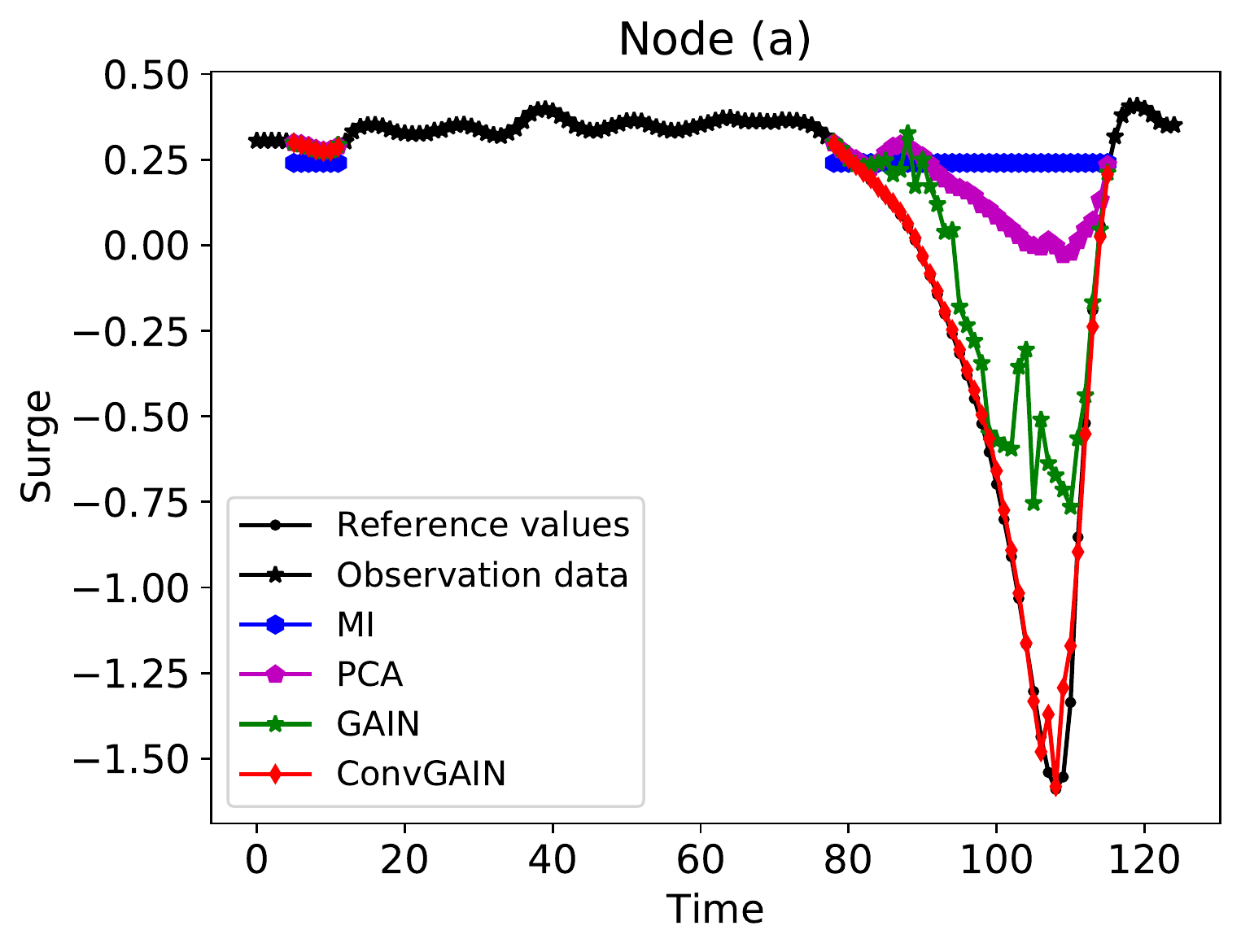}  
      \includegraphics[width=2.5in]{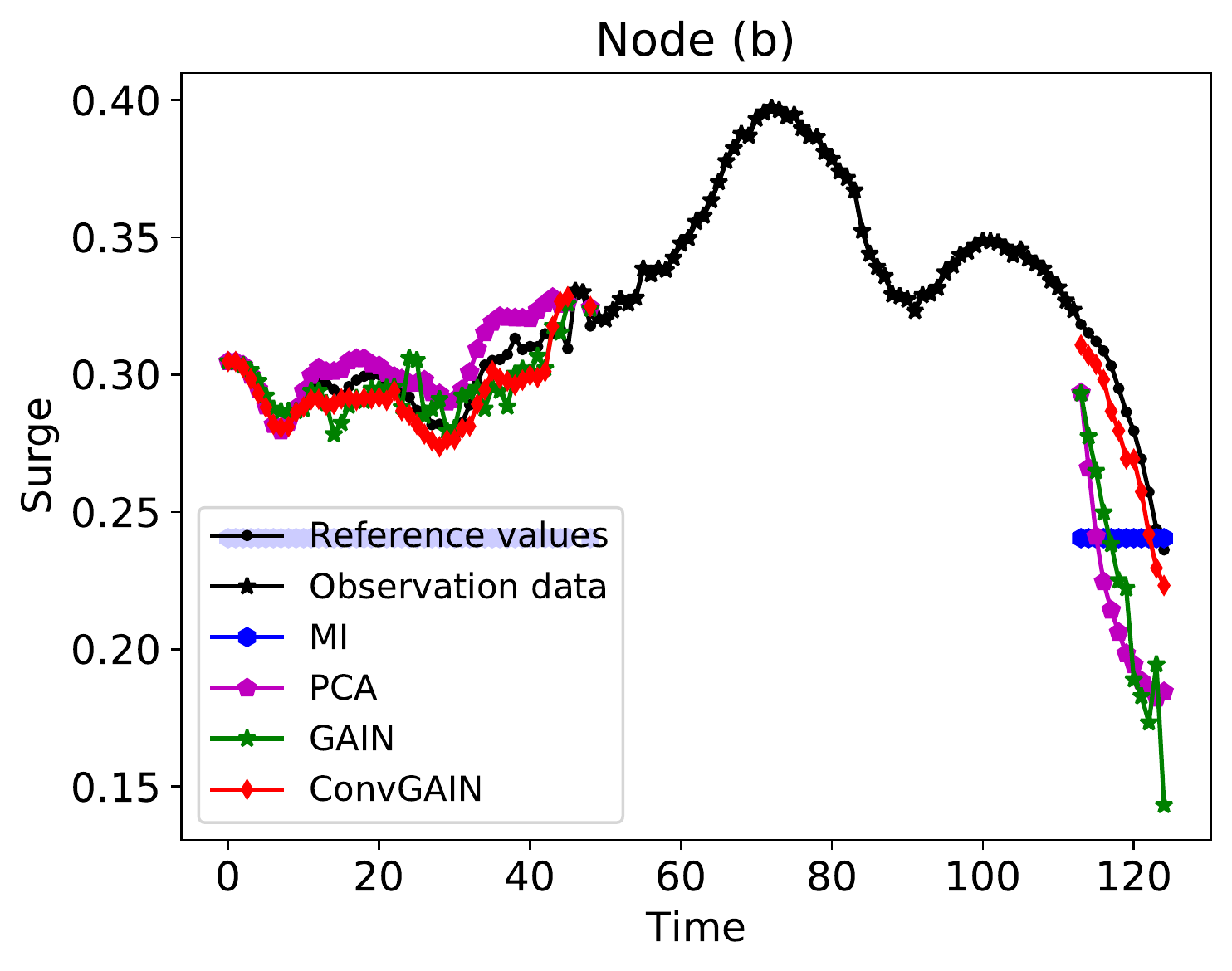}   
      \includegraphics[width=2.5in]{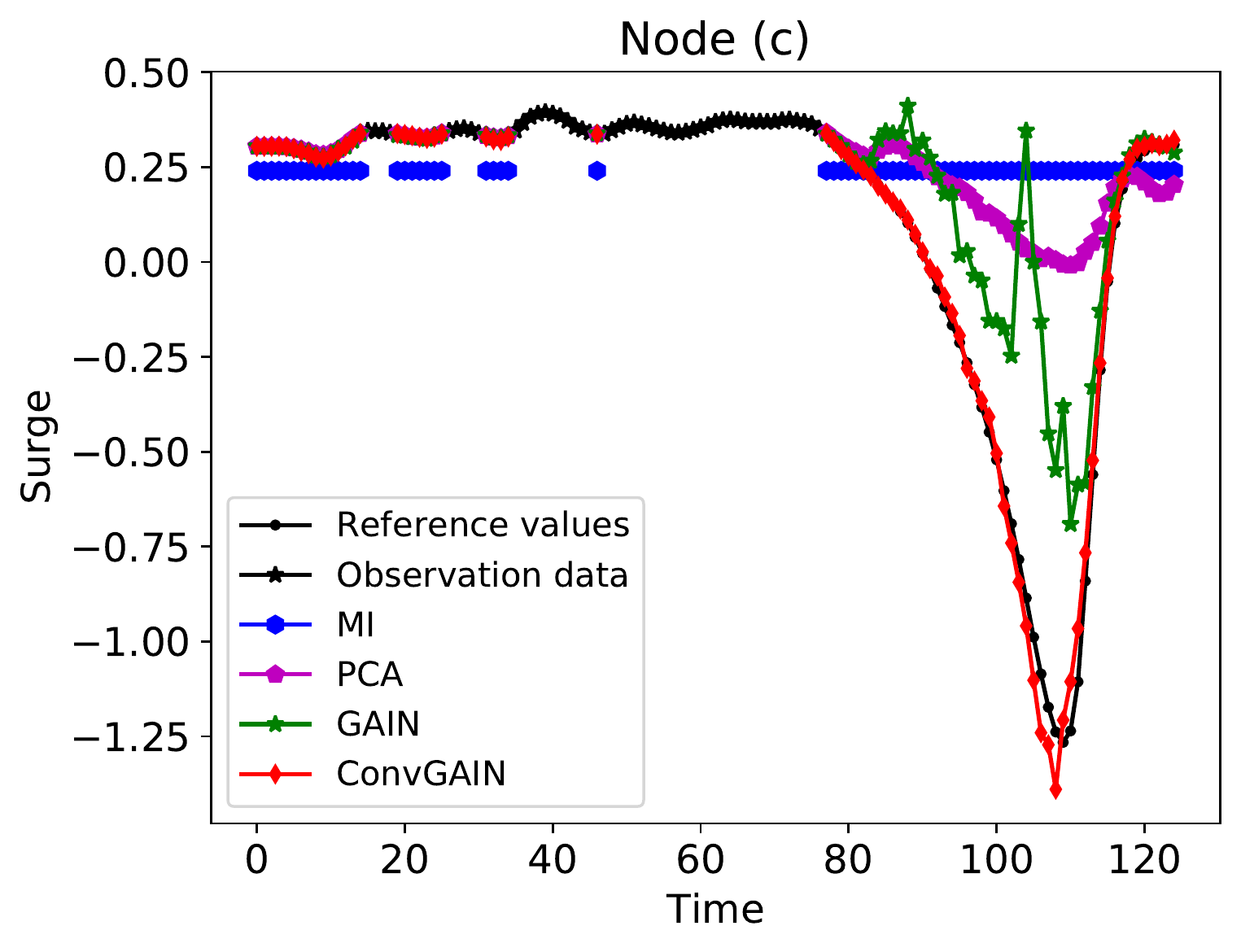}      
      \includegraphics[width=2.5in]{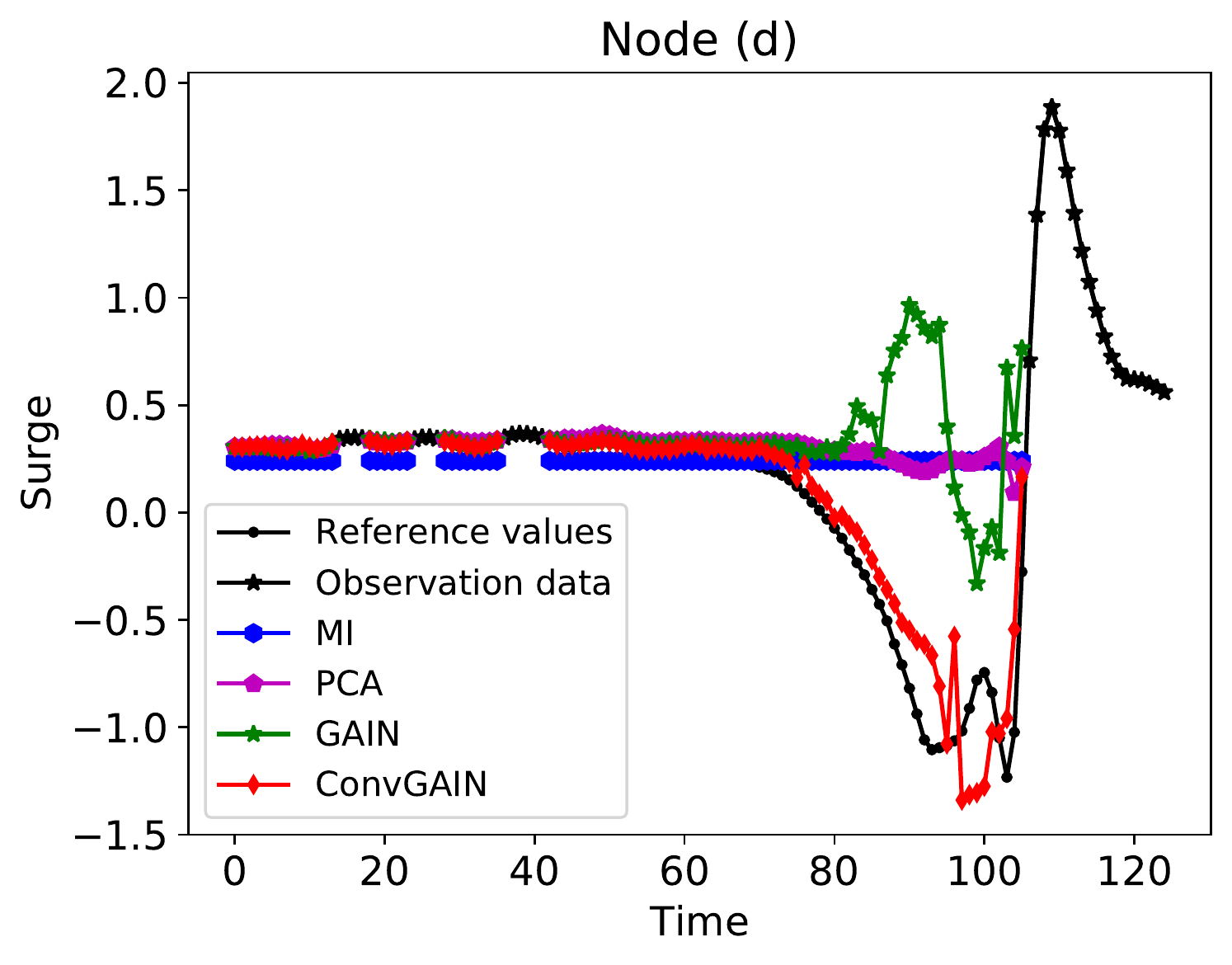}     
      \caption{\label{fig:onenodeimputation} Surge imputation for nodes (a), (b), (c), and (d) with missing rate $36 \%$, $48 \%$, $60 \%$, and $72 \%$ respectively}
    \end{center}\hspace{2pc}%
\end{figure} 

As it is seen in Figure \ref{fig:onenodeimputation}, the Conv-GAIN imputes the missing elements very close to the 
reference values and further provided smoother predictions. Both these are strong indicators that method accommodates higher degrees of learning from the data spatially and temporally. The consequences of improvements and modifications that are applied to the initial GAIN can be clearly seen over the surge imputation for these nodes. 
The GAIN approach can partially predict the surge's behavior, but it does not impute the missing data as effectively as Conv-GAIN. In particular, the GAIN imputation is mostly temporally 
inconsistent with frequent fluctuations, perhaps because GAIN does not learn from the correlation of the data. In comparison to the GAN-based approaches (GAIN and Conv-GAIN), PCA 
offers always over-smoothed predictions that are frequently very far away from the ground truth. 
This is expected, as the simple form of PCA is not expressive enough to learn the imputation pattern in a satisfactory way. 
Performance of MI is even worse as expected and never provided relevant imputation for the missing values.

\subsection{Effect of Structural Distribution of Missing Data and Spatial Features}
\label{subsection:Effect of Structural Distribution of Missing Data and Spatial Features}
As the superiority of Conv-GAIN over other baseline methods is confirmed in the previous section, here we aim to further study the behavior of Conv-GAIN. Specifically, we will investigate how the structural distribution of the missing values can affect the imputation capability of Conv-GAIN, and the benefits of including spatial coordinate features in the input.\\

As a first step to identify the structurally missing data patterns and the corresponding imputation capability of Conv-GAIN, we introduce a process to quantify the degree of structure of the missing data. To illustrate the process, we take two storms both with about thirty percent of total missing data as examples, 
as we observed that the structurally missing data occur more frequently among such datasets with high rate of missing data. 
These storms correspond to the ones have the best and worst structure within the missing data, quantified through the process detailed next, and will be denoted herein as Storm $\text{(a)}$ and Storm $\text{(e)}$.\\ 

The degree of 
structure for the missing patterns 
is quantified here by finding the areas of possible fitted rectangles in the missing subregions of the datasets. We then count the number of rectangles with possible different areas that can be fitted in the missing data regions. The area of the rectangle is defined as the product of consecutive missing values across the temporal and spatial dimensions. Furthermore, we set a cutoff threshold of five (across the temporal dimension) by forty (across the nodal dimension) as the minimal rectangle size to be identified, in order to avoid accounting for overly small missing subregions.
In general, the number of identified rectangles decreases while the given area increases. 
In Figure \ref{fig:rectangles}, we present
the 
characteristics of fitted rectangles given different levels of areas in the missing region for the two aforementioned storms.

\begin{figure}[H]
    \begin{center}{}
      \includegraphics[width=3.0in]{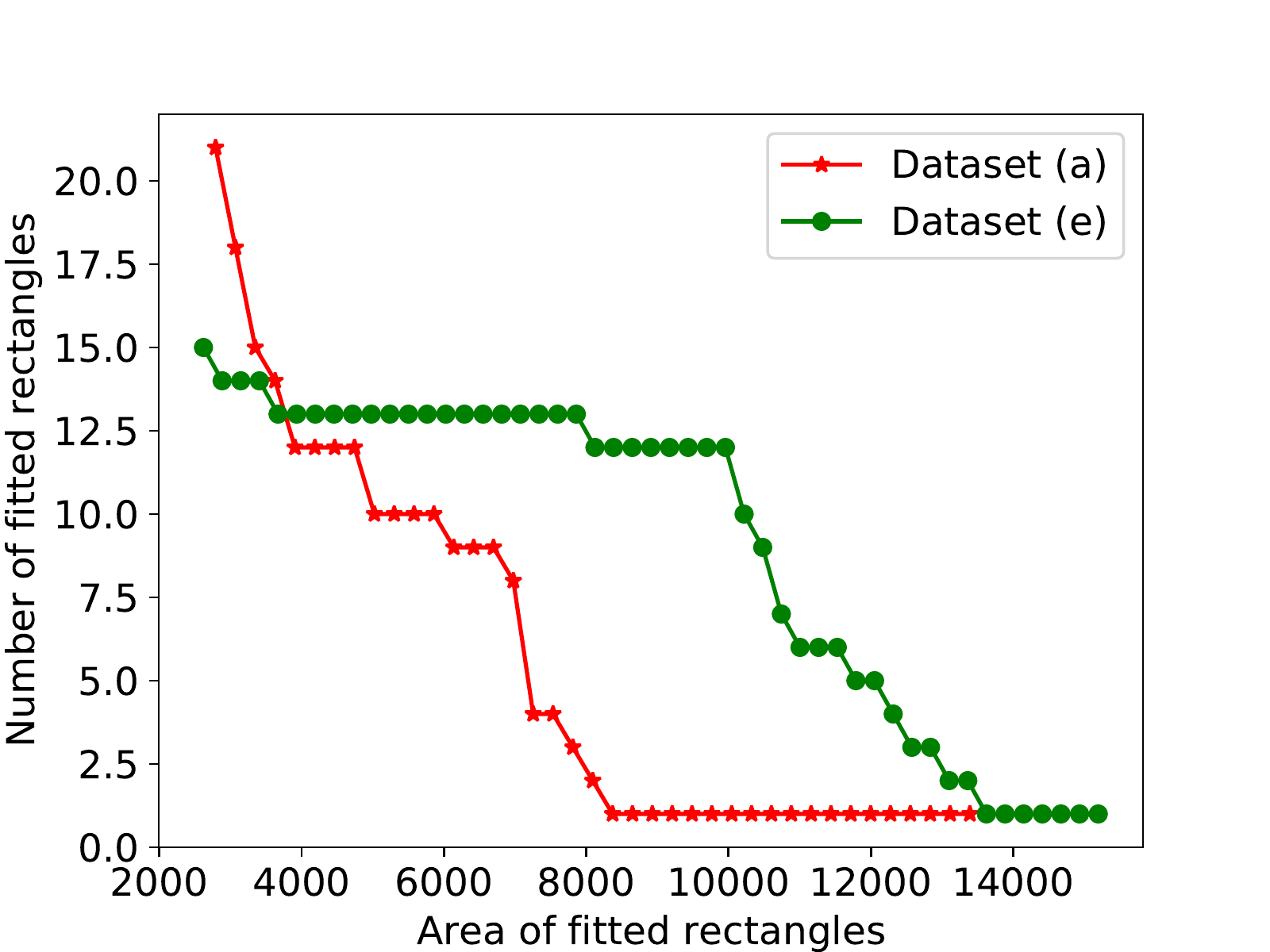}
      \caption {\label{fig:rectangles} Characteristics of fitted rectangles for the missing data}
    \end{center}\hspace{2pc}%
\end{figure} 
As it is seen in Figure \ref{fig:rectangles}, given the same missing area, storm $\text{(e)}$ almost always has higher numbers of fitted rectangles than storm $\text{(a)}$. This indicates that for this storm, we are mainly dealing with blocks, or structurally missing components, rather than uniformly (randomly) missing components. Looking into the imputation performance for both storms, the total RMSE of the storm $\text{(e)}$ $(3.933\mathrm{e}-1)$ is much higher than the RMSE of the storm $\text{(a)}$ $(3.450\mathrm{e}-2)$, suggesting that it is more challenging for Conv-GAIN to deal with storm datasets with structurally missing data.\\

To further confirm our observation, the same procedure is applied 
for all five storms corresponding to the $30\%$ missing rate.
Figure \ref{fig:rectangles5} similarly to Figure \ref{fig:rectangles} shows the 
characteristics of fitted rectangles with different areas. For the sake of illustration clarity, we increase the cutoff of the minimal fitted rectangle when plotting
for these five datasets to a total area of fitted rectangles 6000 squared units, as the larger blocks of missing components are the much harder regions to impute in comparison to small blocks of missing components. This allows us to focus to bigger size gaps for the missing data, compared to the earlier examination in Figure \ref{fig:rectangles}.\\

\begin{figure}[H]
    \begin{center}{}
      \includegraphics[width=3.0in]{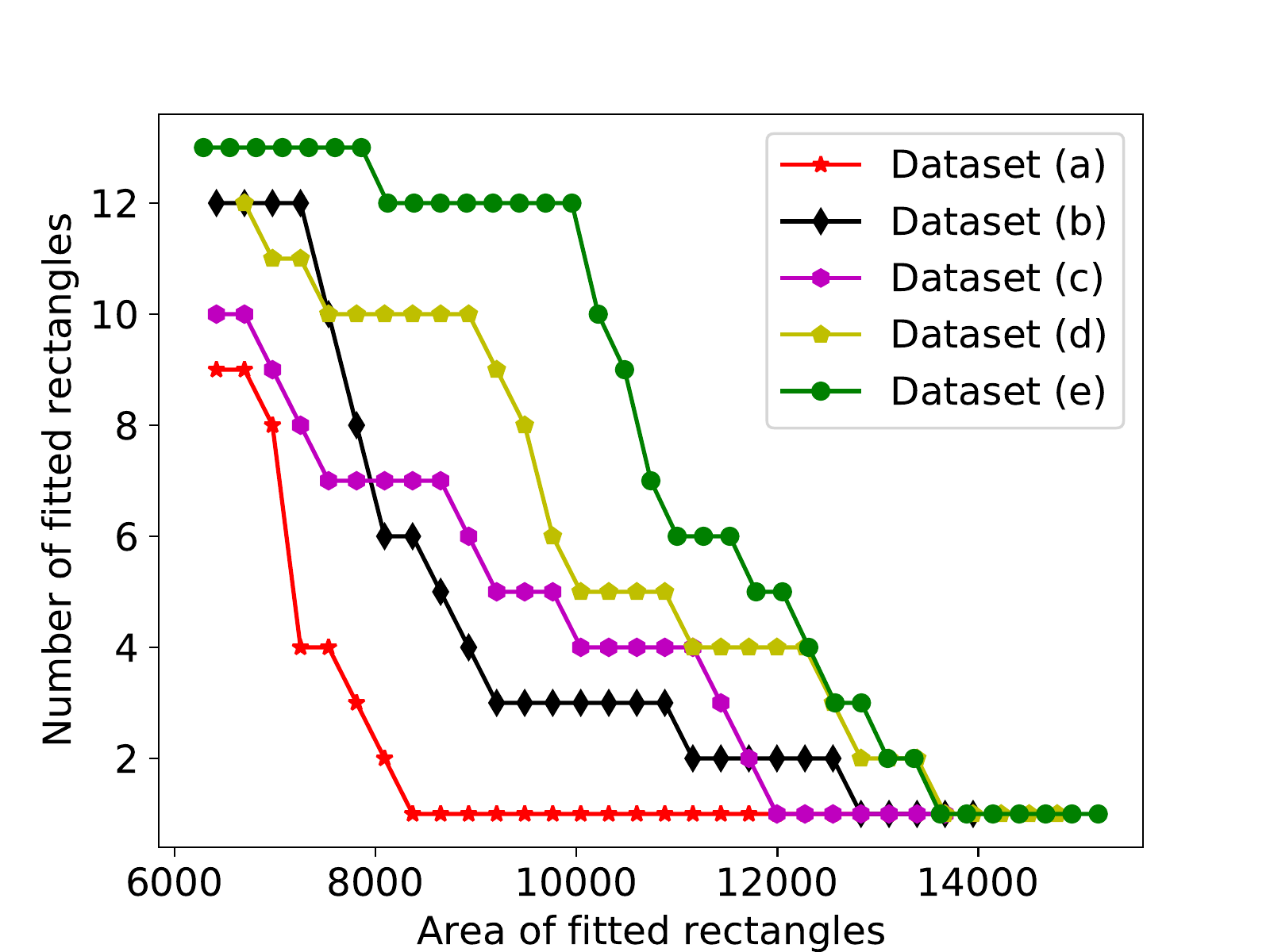}
      \caption {\label{fig:rectangles5} Maximum number of fitted rectangles in the missing subregions}
    \end{center}\hspace{2pc}%
\end{figure} 

Furthermore in this comparison we consider one other variant of our proposed Conv-GAIN approach, to examine the benefit of considering the spatial node coordinates as additional input features. Specifically, we implement Conv-GAIN models while only considering the storm surge data as the input (thus ignore the spatial coordinates). We denote this approach as Conv-GAIN w/o coordinates. We also obtain the performance of the proposed Conv-GAIN and the baseline approach GAIN, and present the RMSEs of all approaches on the datasets $(a)$ to $(e)$ in Table \ref{tab:RMSE5}.

\begin{table}[H]
\caption{RMSE of missing data}
\vspace{-0.5cm}
\begin{center}
\scalebox{0.8}{
\begin{tabular}{l*{7}{c}r }
\hline
Dataset      & $\text{(a)}$  &  $\text{(b)}$ & $\text{(c)}$ & $\text{(d)}$ &$\text{(e)}$ &\\
\hline
  $\text{GAIN}$                  & $2.026\mathrm{e}-1$ & $1.929\mathrm{e}-1$ & $1.688\mathrm{e}-1$ &  $3.574\mathrm{e}-1$ & $4.665\mathrm{e}-1$  & \\
  $\text{\bf{Conv-GAIN}}$                  & $\bf{3.450\mathrm{e}-2}$ & $\bf{1.309\mathrm{e}-1}$ & $\bf{1.432\mathrm{e}-1}$ &  $\bf{3.518\mathrm{e}-1}$ & $\bf{3.933\mathrm{e}-1}$  & \\
  $\text{Conv-GAIN w/o} $                 & $4.923\mathrm{e}-2$ & $1.603\mathrm{e}-1$ & $1.682\mathrm{e}-1$ &  $3.561\mathrm{e}-1$ & 
  $4.281\mathrm{e}-1$  & \\
  $\text{coordinates} $\\
\hline
\end{tabular}}\label{tab:RMSE5}
\end{center}
\end{table}

Comparing the RMSEs in Table \ref{tab:RMSE5}, 
it can be inferred that for both Conv-GAIN approaches, the RMSEs increase from dataset $(a)$ to $(e)$ as the datasets contain more structurally missing data (shown in Figure \ref{fig:rectangles5}). This indicates that besides the total missing rate of data, the structure of missing data can also strongly impact the imputation performance of Conv-GAIN. Interestingly, looking at the first row, we do not observe strong dependency between the imputation errors with the degree of data being structurally missing for GAIN.\\

Moving to the comparison among different approaches, we observe that for all datasets, Conv-GAIN performed the best, followed by Conv-GAIN w/o coordinates and GAIN. These results show that considering the coordinates enriches the provided information to the model and results in a better imputation performance. Even the sole use of convolutional layers can bring considerable performance gain over the baseline imputation methods (compare Conv-GAIN w/o coordinates with GAIN).\\

Finally, it should be pointed out that the particular way the storm surge missing data are structurally distributed was the main reason for developing Conv-GAIN. As we observe from the experimental result, Conv-GAIN can urge the neural network model to learn from the neighborings' regions of the unobserved blocks (structurally missing data), and extract meaningful information from the spatial and temporal correlation of data. In the comparisons examined in this subsection, we also showed that considering the coordinates of nodes as additional features could also improve the performance of Conv-GAIN. The results also further confirm that Conv-GAIN provides better informed imputation process than the original GAIN, especially for the type of corrupted datasets like the ones examined in the storm surge simulations examined here.

\section {Conclusion}
\label{section:Conclusion}
In this work, a new GAN-based method is introduced to impute the missing data. The developed method named Convolutional Generative Adversarial Imputation Net (Conv-GAIN) is a combination of Convolutional Neural Networks (CNN) and Generative Adversarial Imputation Nets (GAIN). The motivation of the method is to impute the spatio-temporal missing data existing in 
storm surge simulation datasets, though the established advances can be employed to other problems with similar missing data characteristics. Employing Conv-GAIN, we are able to learn the correlation of data spatially and temporally 
by learning from the adjacent elements of data.
Furthermore, the coordinates of nodes as additional features are used to provide Conv-GAIN structure with more information. The performance of Conv-GAIN on the studied corrupted data shows that capturing the correlation through convolutional layers results in better imputation of missing data compared to the GAIN where fully connected layers are only used. This difference of Conv-GAIN and GAIN performances is more significant for data sets with higher total rates of missing data. The proposed method should be suggested for 
datasets 
with significant underlying data correlation, 
especially for datasets 
for which the missing subregions are structurally distributed in blocks of different sized at random. It is important to acknowledge that for datasets with higher number of larger subregions of missing data the performance of Conv-GAIN deteriorates, as 
as correlation to neighbors without missing values decreases, making learning through convolutional layers harder.\\

\noindent \textbf{\textit{Acknowledgement}} \\
Authors would like to thank the Army Corp of Engineers, Coastal Hydraulics Laboratory of the Engineering Research and Development Center for providing access to the storm surge data that were used in the illustrative case study. 
%

\newpage


\appendix
\renewcommand{\thesection}{\Alph{section}.\arabic{section}}
\setcounter{section}{0}

\begin{appendices}

\section{Dataset Characteristics}
\label{Dataset Characteristics}

The surge heat-map for the storms considered in the illustrative application is shown in Figure \ref{fig:stormsmissingregions}. Storms are grouped according to their missing rate, with missing components indicated in white color. It should be pointed out that each dataset has a different range of surge values, 
so the 
heat-map colors are not consistent across them. Since emphasis is on the structure or missing data, and not necessarily the surge values themselves, the colorbar ranges for each storm are not reported.

\begin{figure}[H]
    \begin{center}{}
      \includegraphics[width=5in]{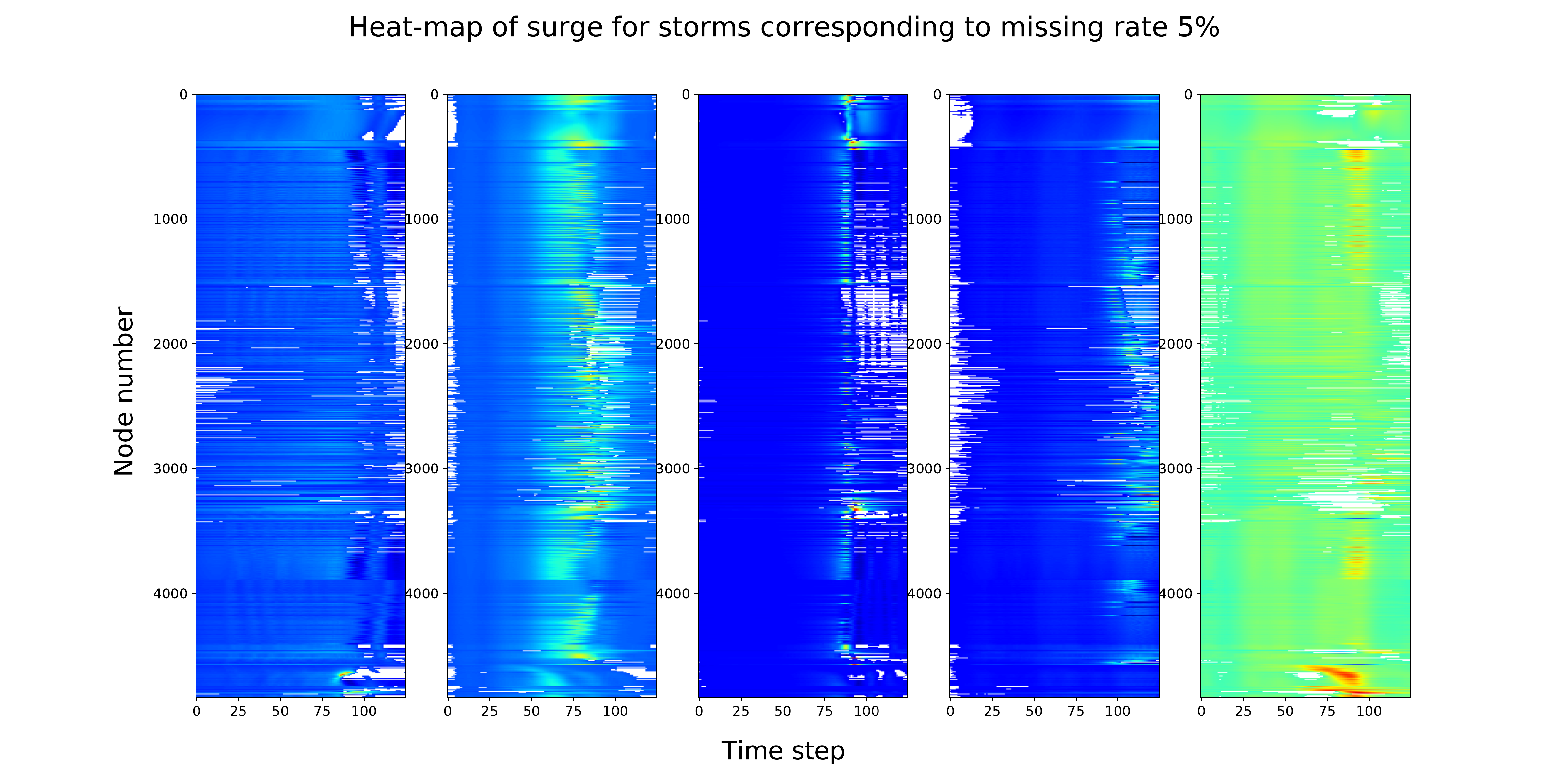}
    \end{center}\hspace{2pc}%
\end{figure} 

\vspace{-2cm}

\begin{figure}[H]
    \begin{center}{}
      \includegraphics[width=5in]{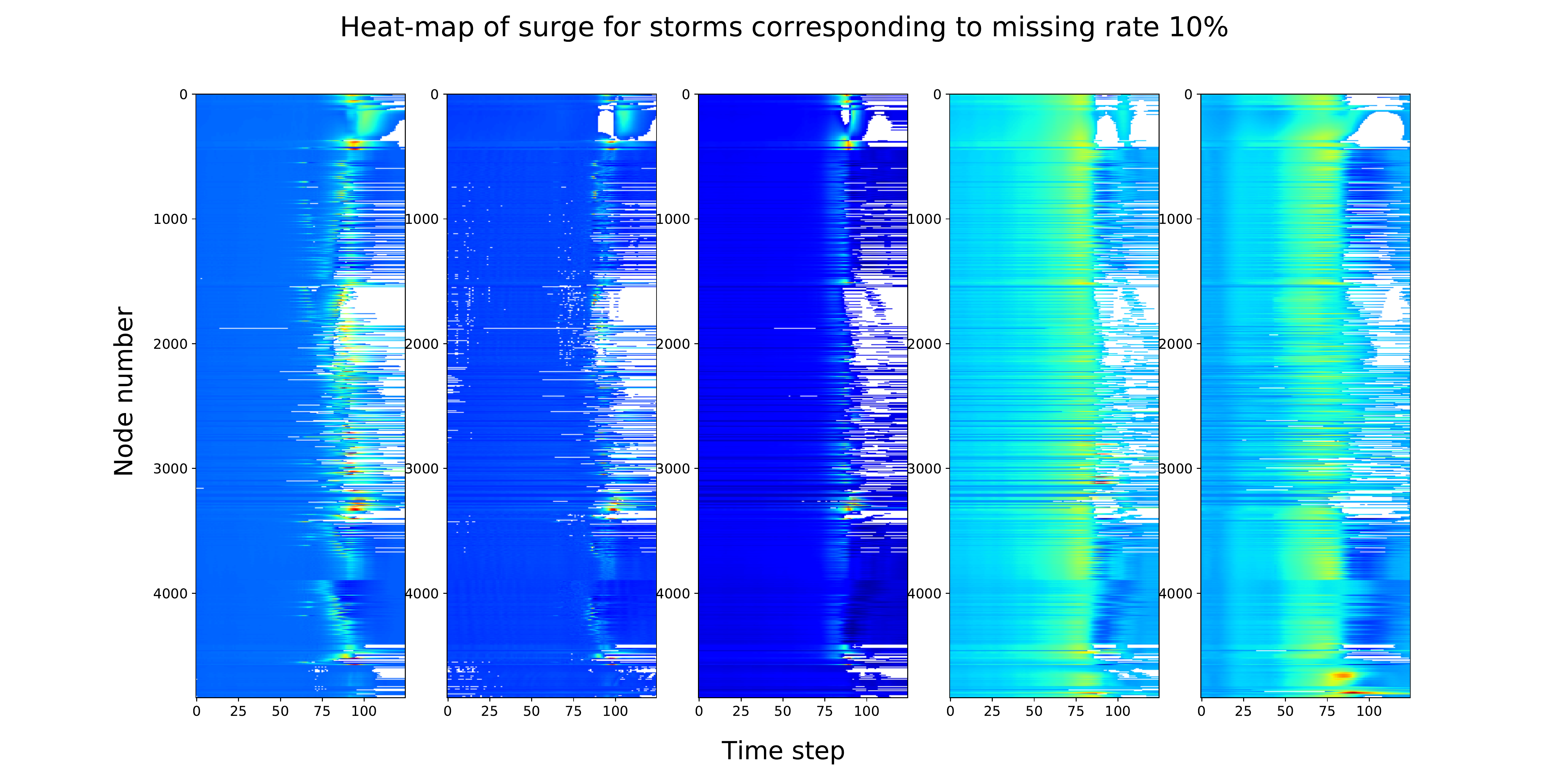} 
    \end{center}\hspace{2pc}%
\end{figure} 

\vspace{-2cm}

\begin{figure}[H]
    \begin{center}{}
      \includegraphics[width=5in]{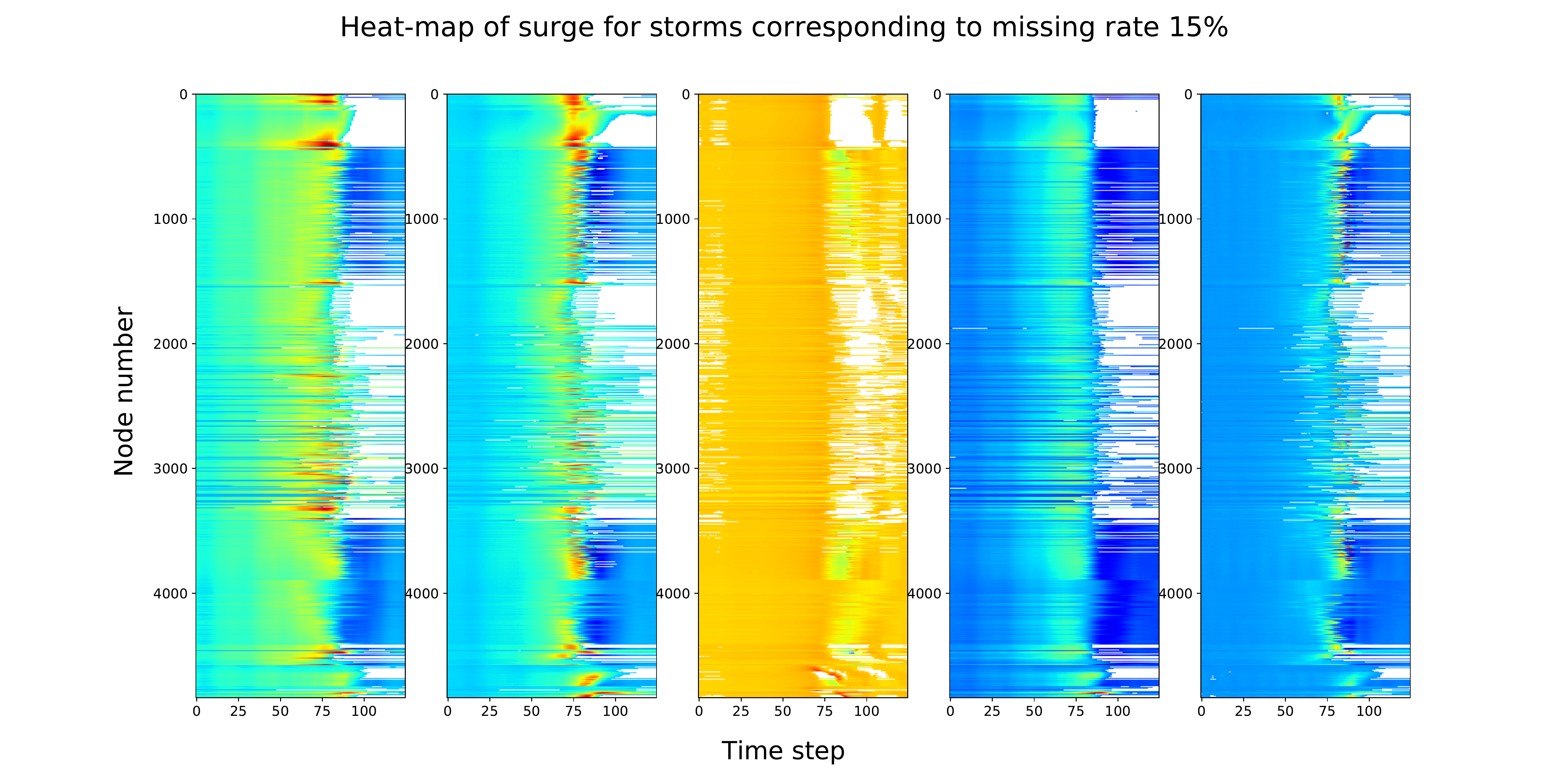} 
    \end{center}\hspace{2pc}%
\end{figure} 

\vspace{-2cm}

\begin{figure}[H]
    \begin{center}{}
      \includegraphics[width=5in]{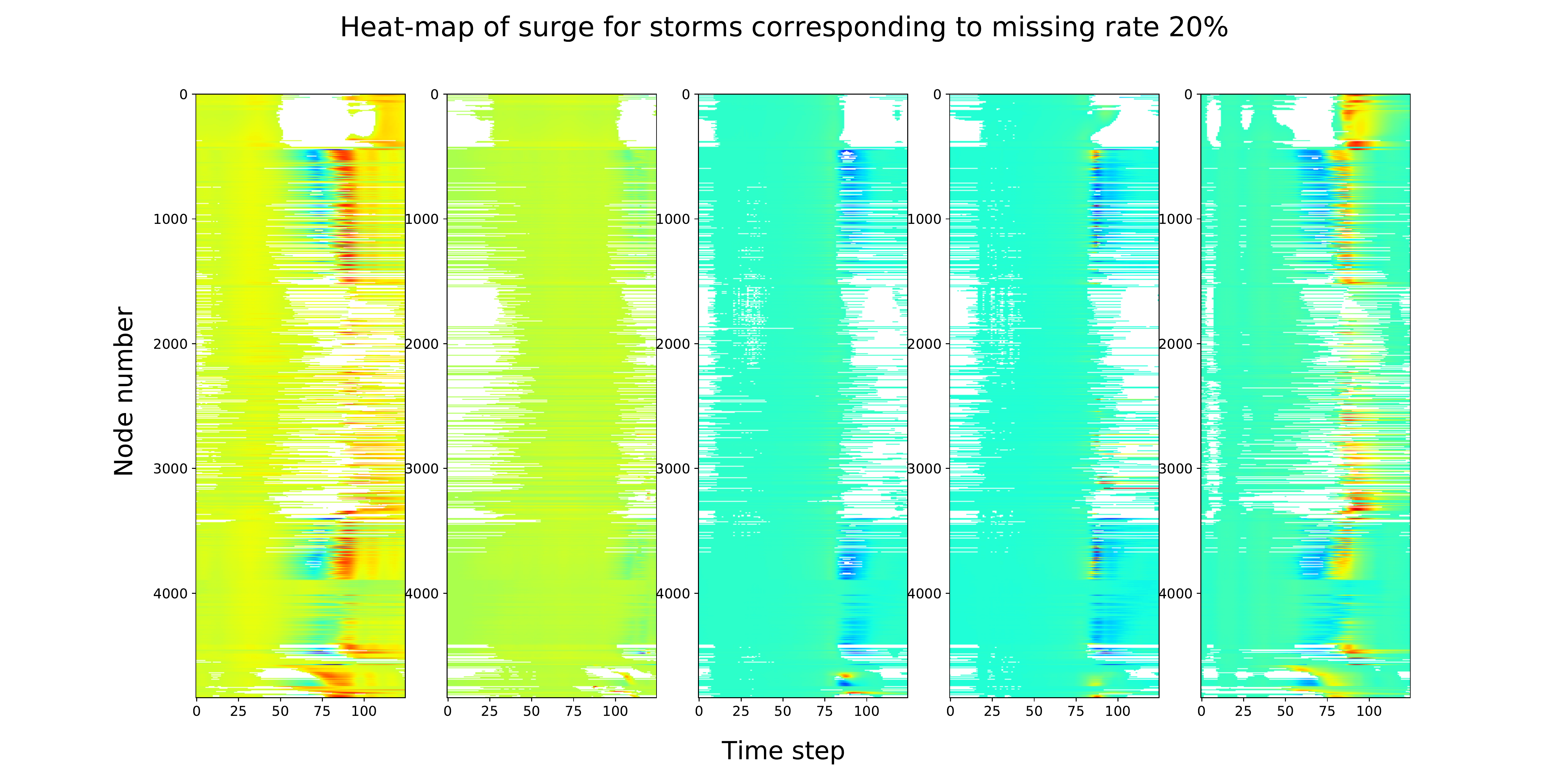} 
    \end{center}\hspace{2pc}%
\end{figure} 

\vspace{-2cm}

\begin{figure}[H]
    \begin{center}{}
      \includegraphics[width=5in]{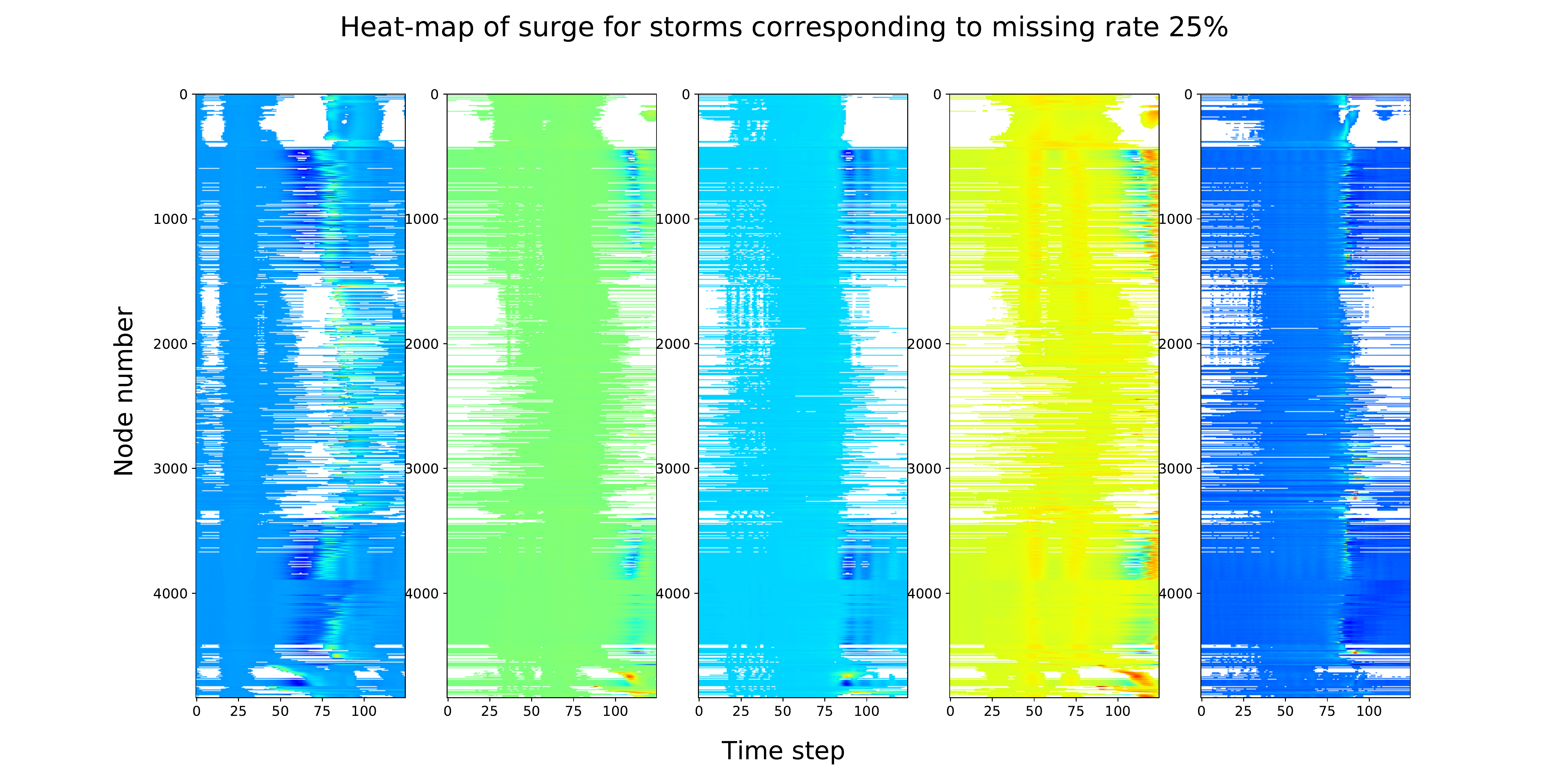} 
    \end{center}\hspace{2pc}%
\end{figure} 

\vspace{-2cm}

\begin{figure}[H]
    \begin{center}{}
      \includegraphics[width=5in]{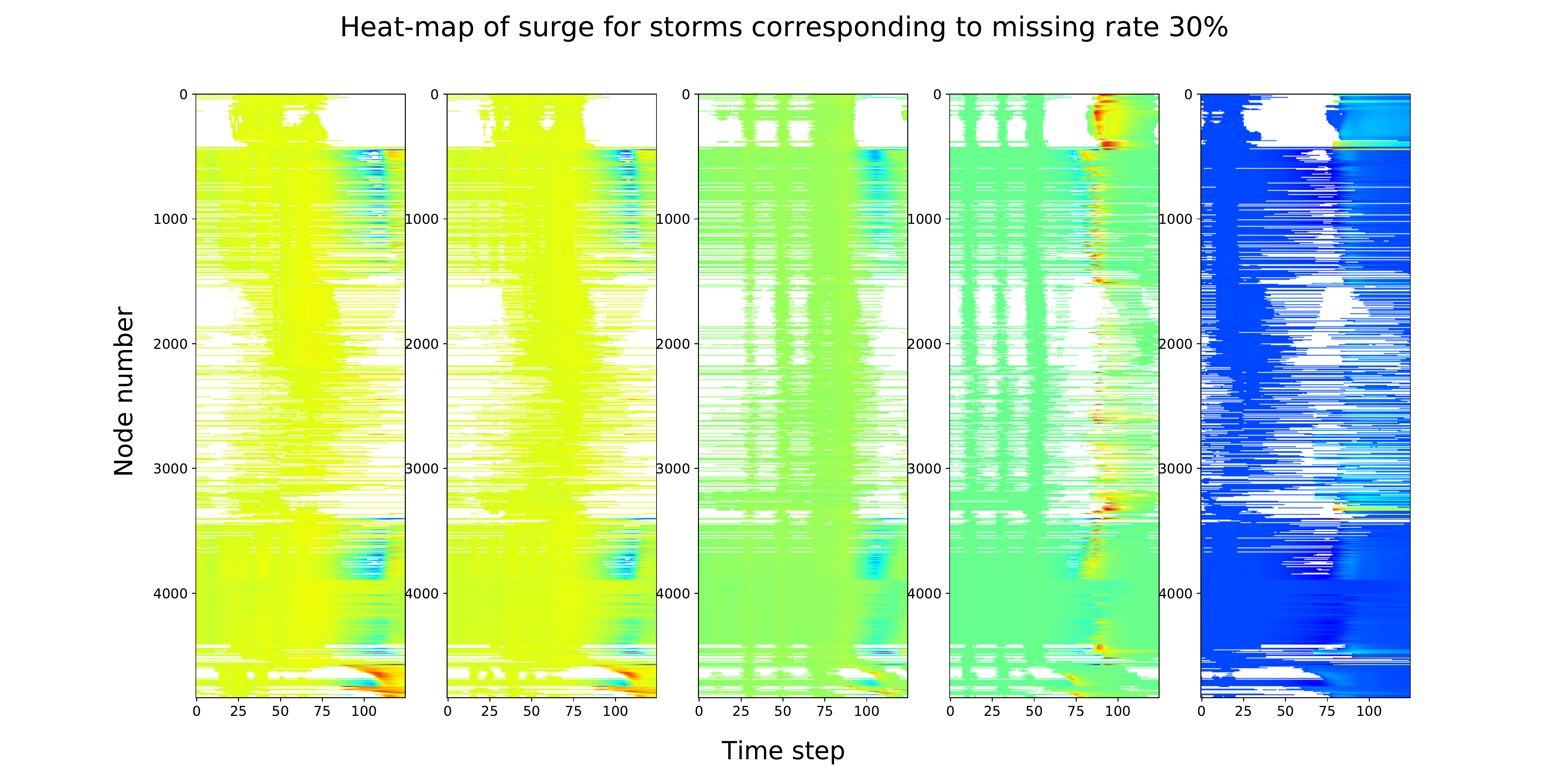} 
      \caption{\label{fig:stormsmissingregions} Heat-map of the storm surge for the datasets considered in the illustrative application. Note that the surge range for the heat map is different per storm, adjusted to the minimum and maximum observed values for that storm (so colors are not consistent across the storms).}
    \end{center}\hspace{2pc}%
\end{figure} 

\end{appendices}

\end{document}

%% file: components/settings-commands.tex
\usepackage{amsfonts}
\usepackage{amsmath}
\usepackage{enumerate}

%% file: components/iformula.tex
\newcommand{\ignore}[1]{}

\newcommand{\vek}[1]{\mathchoice{\displaystyle\boldsymbol{#1}}
{\textstyle\boldsymbol{#1}}{\scriptstyle\boldsymbol{#1}}
{\scriptscriptstyle\boldsymbol{#1}}}

\newcommand{\mat}[1]{\mathchoice{\displaystyle\mathbf{#1}}
{\textstyle\mathbf{#1}}{\scriptstyle\mathbf{#1}}
{\scriptscriptstyle\mathbf{#1}}}

\newcommand{\ops}[1]{\mathchoice{\displaystyle\mathsf{#1}}
{\textstyle\mathsf{#1}}{\scriptstyle\mathsf{#1}}
{\scriptscriptstyle\mathsf{#1}}}

